\documentclass[letterpaper, 10 pt, journal]{ieeetran}  

\IEEEoverridecommandlockouts                              

\usepackage{graphicx}
\graphicspath{{figs/}}
\usepackage{amsmath}
\usepackage{amssymb}
\usepackage{amsfonts}
\usepackage{algorithm}
\usepackage{algpseudocode}
\usepackage{bm}
\usepackage{breqn}
\usepackage{color}
\usepackage{gensymb}
\usepackage{here}
\usepackage{physics}
\usepackage{siunitx}
\usepackage{times}
\usepackage{url}
\newcommand{\figref}[1]{Fig.~\ref{figure:#1}}
\newcommand{\tabref}[1]{Table~\ref{table:#1}}
\newcommand{\secref}[1]{Sec.~\ref{section:#1}}

\title{\LARGE \bf
Design and Control of a Small Humanoid Equipped with\\Flight Unit and Wheels for Multimodal Locomotion
}

\author{Kazuki Sugihara, Moju Zhao, Takuzumi Nishio, Tasuku Makabe, Kei Okada, and Masayuki Inaba%
}

\begin{document}

\maketitle


\begin{abstract}
Humanoids are versatile robotic platforms owing to their limbs with multiple degrees of freedom.
Although humanoids can walk like humans, they are relatively slow, and cannot run over large barriers.
To address these limitations, we aim to achieve rapid terrestrial locomotion ability and simultaneously expand the locomotion domain to the air by utilizing thrust for propulsion.
In this paper, we first describe an optimized construction method for a humanoid robot equipped with wheels and a flight unit to achieve these abilities.
Then, we describe the integrated control framework of the proposed flying humanoid for each locomotion mode: aerial, legged, and wheeled locomotion.
Finally, we achieved multimodal locomotion and aerial manipulation experiments using the proposed robot platform.
To the best of our knowledge, this is the first time that a single humanoid has simultaneously achieved three different types of locomotion, including flight.


\end{abstract}

\section{INTRODUCTION}
\IEEEPARstart{H}{umanoid} robots are robot platforms with multi-degree-of-freedom limbs, which allow them to perform several tasks like humans.
Consequently, several studies have been conducted on the application of humanoids in human living spaces \cite{Kaneko2004hrp2} and disaster sites \cite{JAXON:Kojima:Humanoids2015}.
Accordingly, various locomotion methods for humanoids have been developed \cite{UnifiedBalance:Kojio:IROS2019}\cite{Kajita2003preview}\cite{Kajita2010invpend}.
However, achieving high speed locomotion by bipedal walking is relatively challenging compared to other terrestrial locomotion types like wheeled locomotion because the support polygon area for bipedal robots, such as a full-body humanoid, is generally small and thus it is easy to lose balance, especially in uneven terrain.
Another challenge to bipedal walking is that its domain of locomotion is limited to the ground.
In environments with barriers that cannot be overcome by bipedal limbs, locomotion becomes difficult.
It is important to expand the locomotion domain of humanoids  to perform tasks in diverse environments.
In particular, flying is effective for overcoming uneven or steep terrain.

\begin{figure}[t]
  \centering
  \includegraphics[width=1.0\columnwidth]{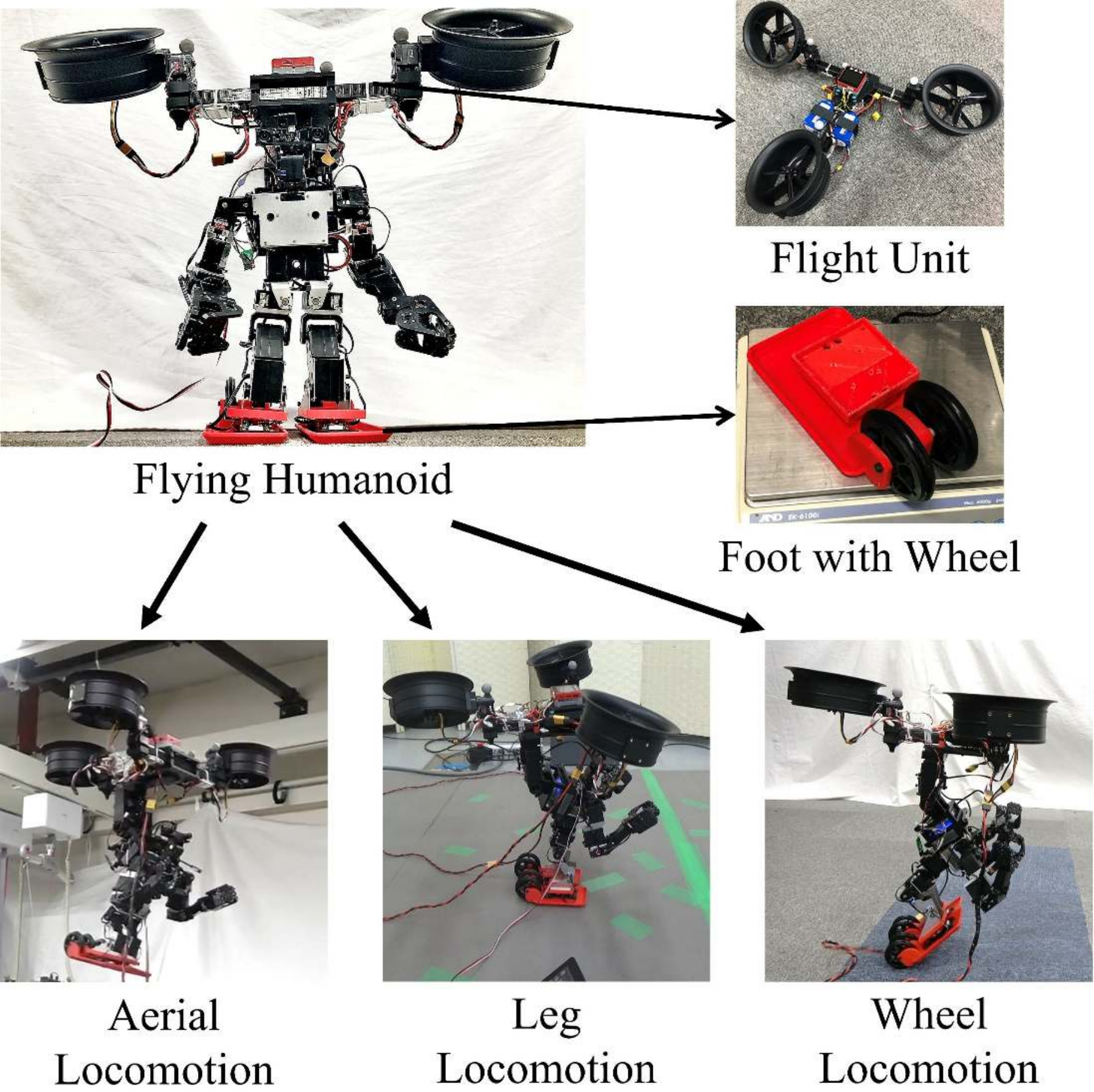}
  \vspace{-6mm}
  \caption{\textbf{Proposed flying humanoid}. It equipped with wheels on foot and a fully-actuated trirotor flight unit. Multimodal locomotion is achieved with this robot platform. Lower left: aerial locomotion using thrust. Lower center: legged locomotion. Lower right: wheeled locomotion using thrust and wheels deployed on the foot.}
  \label{figure:overview}
  \vspace{-5mm}
\end{figure}

Having multiple locomotion methods and selecting the appropriate alternative can facilitate easy adaptation to diverse environments and saving in energy and time consumption.
Therefore, in this work, we aim to achieve rapid terrestrial and aerial locomotion by humanoids, as shown in \figref{overview}.

Several humanoids exist with terrestrial locomotion mechanisms other than bipedal walking such as wheels \cite{Jung2018hubo} or crawlers \cite{Haynes2017chimp}.
Compared to walking by bipedal legs, these locomotion methods are based on wheel rotational motion, which makes it easier to achieve high-speed locomotion.
In \cite{Page2014uavugv}, wheels were deployed to a drone for terrestrial locomotion by aerial robots.
Thrust can be applied to stabilize the motion of the humanoid presented in \cite{huang2017jet} and enables aerial locomotion.
Therefore, we propose a humanoid equipped with wheels and a flight unit to achieve rapid terrestrial and aerial locomotion.
Regarding the humanoid presented in \cite{Jung2018hubo} and \cite{Haynes2017chimp}, the employed wheels and crawlers required additional actuators to drive them\; however, an increase in weight due to these attached mechanisms severely limits aerial locomotion.
Hence, we propose the adoption of passive wheels, which enable a minimum increase in weight and utilize thrust for propulsion.

Some robots have legs and thrusters attached to their torso for aerial locomotion.
In \cite{kim2021bipedal}, a bipedal robot that combines a quadrotor was developed.
Although this robot can move through multiple environments, it does not have manipulators.
Full-body flying humanoid platform presented in \cite{anzai2021design} with a birotor flight unit could walk and fly.
The birotor and quadrotor in these related works are under-actuated, which causes problems such as a lack of full pose trackability in aerial locomotion.
Therefore, we deploy a flight unit with fully-actuated flight control to achieve full pose trackability for a flying humanoid.
By increasing the number of rotors, we can increase the number of control degrees of freedom.
However, if the number of rotors is excessively increased, the humanoid's motion may become unstable owing to the resulting high center of gravity (CoG) and this approach may cause aerodynamic interference from rotors on arms.
Therefore, we employed a thrust vectoring mechanism to realize fully-actuated flight control with three rotors.
The thrust vectoring mechanism provides a degree of freedom (DOF) to tilt the rotor.
Using the thrust vectoring mechanism, the control DOF is increased and it is employed to achieve control redundancy \cite{DRAGON:Chou:ICRA2018_RAL} or exert forces to the environment \cite{DroneDoorOpening:Sugito:ICRA2022}.
This mechanism enables flying with less number of rotors such as 
birotors \cite{Qin2020bicopter} or trirotors \cite{Hu2018trirotor}.


In addition to flying, humanoids with thrusters can use thrust to stabilize their motion.
In \cite{huang2017jet}, a model based thrust control was adopted to assist motion and overcome barriers using a bipedal robot with thrusters.
Reinforcement learning was adopted to calculate thrust output to stabilize walking motion against disturbance in \cite{Shi2022learning}.
Thrust was also employed to reduce the load torque on the leg joints of a quadruped robot with thrusters \cite{zhao2023spidar}.
In this work, an integrated control framework is proposed for flying humanoids to achieve stabilization, maneuverability, and a decrease in the load torque on joints.
This control framework can switch locomotion mode seamlessly.

The main contribution of this work is summarized as follows:
\begin{enumerate}
  \item We propose an optimized construction method for a humanoid robot equipped with wheels and a flight unit to achieve rapid terrestrial and aerial locomotion.
  \item We present an integrated control framework for the proposed flying humanoid with three modes of locomotion: aerial, legged, and wheeled locomotion.
  \item We achieve multimodal locomotion and aerial manipulation by the proposed flying humanoid robot.
\end{enumerate}

The remainder of this paper is organized as follows.
The design and modeling of the flying humanoid are introduced in \secref{2}.
Based on the optimization of orientation, the desired configuration of the flying humanoid is also presented in \secref{2}.
In \secref{3}, an integrated control framework for the proposed flying humanoid is described.
Subsequently, we demonstrate the prototype of the flying humanoid and present experimental results in \secref{experiment}.
Finally, conclusions are presented in \secref{conclusion}.

\section{DESIGN AND MODELING}
\label{section:2}

\subsection{Flight Unit}
In this work, a fully-actuated trirotor was deployed on the proposed humanoid.
This flight unit contains three rotors and a DOF to tilt each rotor.
By properly arranging three rotors that can exert forces in two directions, it is possible to realize fully-actuated control that enables control position and attitude independently.
This makes it possible to exert torque on the humanoid without exerting translational forces.

\begin{figure}
  \centering
  \includegraphics[width=0.8\columnwidth]{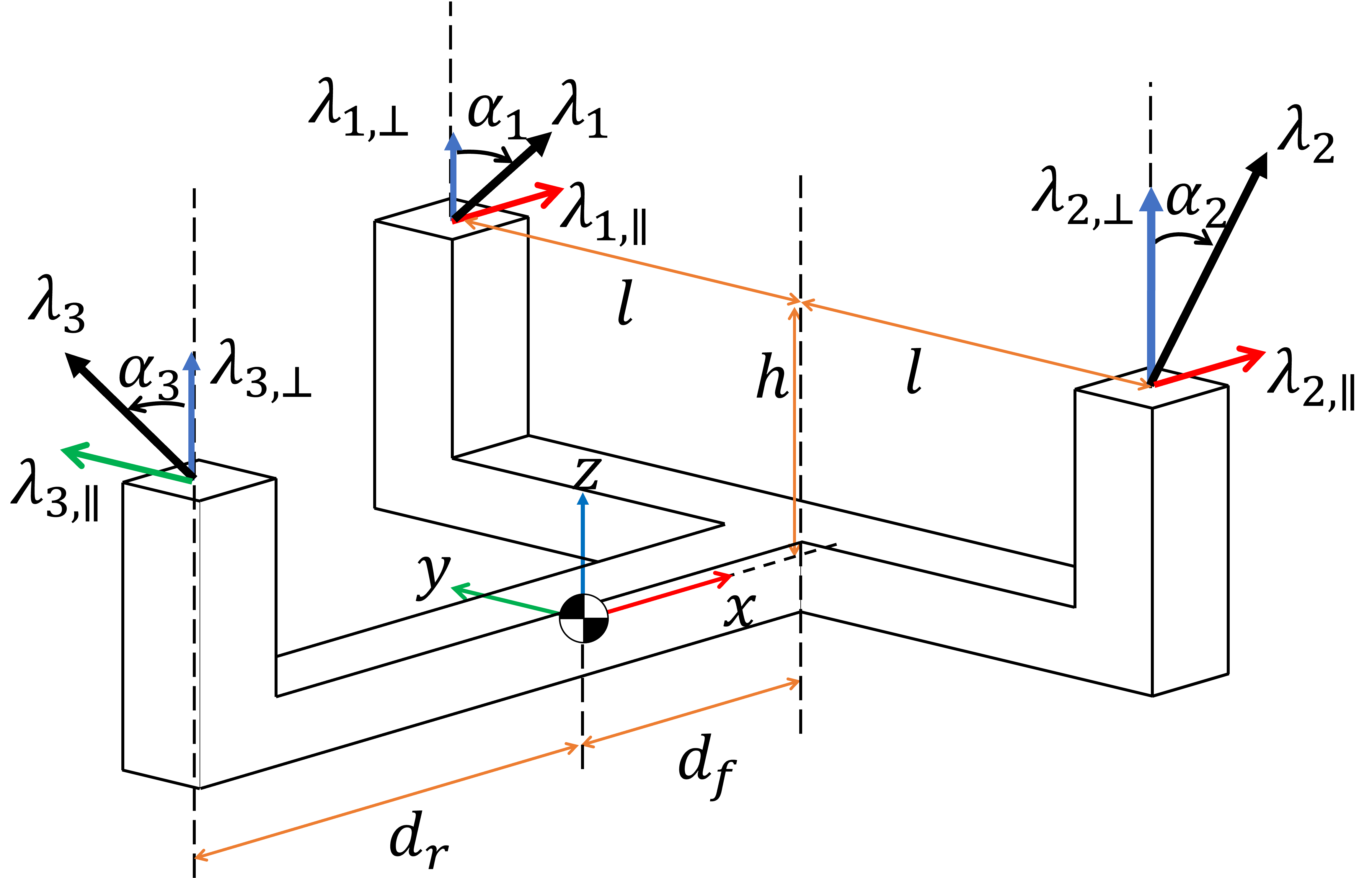}
  \caption{\textbf{Dynamics model of fully-actuated trirotor flight unit}. Each rotor has vectoring freedom so each rotor can exert force in two directions.}
  \label{figure:trirotor dynamics}
  \vspace{-5mm}
\end{figure}

The motions of the joints affect the position of CoG or flight stability; however, we assumed that motions of the joints are slow, and applied the quasi-static assumption. 
Moreover, changes in robot model due to joint motion are compensated by integral control terms in flight control.
The dynamics model of the fully-actuated trirotor in this work is depicted in \figref{trirotor dynamics}.
From the Newton-Euler equations, translational and rotational motions are described as follows respectively
\begin{gather}
  M \ddot{\bm{r}} = -M \bm{g} + \bm{R} \bm{f},\\
  \bm{I} \dot{\bm{\omega}}_{\lbrace CoG \rbrace} = \bm{\tau} - \bm{\omega}_{\lbrace CoG \rbrace} \cross \bm{I} \bm{\omega}_{\lbrace CoG \rbrace},
\end{gather}
where \(M\) denotes the mass of the robot. 
In addition, \(\bm{r}\), \(\bm{g}\), \(\bm{\omega}_{\lbrace CoG \rbrace}\), \(\bm{f}\), \(\bm{\tau}\) \(\in \mathbb{R}^{3}\) represent the position of the CoG frame w.r.t  the world frame, gravitational vector, angular velocity of the CoG frame, force and torque generated by the thrusts in the CoG frame, respectively.
\(\bm{R}\), \(\bm{I}\) \(\in \mathbb{R}^{3 \times 3}\) denote the rotation matrix of the CoG frame w.r.t the world frame and inertia matrix of the body, respectively.
\(\bm{f}\) and \(\bm{\tau}\) are described in the CoG frame.
From \figref{trirotor dynamics}, the force and torque in the CoG frame by vectoring thrust force \(\bm{\lambda} \in \mathbb{R}^{6}\) can be described as follows using \(\bm{Q} \in \mathbb{R}^{6 \times 6}\)
\begin{equation}
  \begin{bmatrix}
    \bm{f}\\\bm{\tau}
  \end{bmatrix}
  =
  \bm{Q}
  \begin{bmatrix}
    \lambda_{1, \perp}&
    \lambda_{1, \parallel}&
    \cdots&
    \lambda_{3, \perp}&
    \lambda_{3, \parallel}
  \end{bmatrix}^T \label{wrench allocation}
  =
  \bm{Q}\bm{\lambda},
\end{equation}
\begin{equation}    
  \bm{Q}
  =
  \begin{bmatrix}
    0 & 1 & 0 & 1 & 0 & 0\\
    0 & 0 & 0 & 0 & 0 & 1\\
    1 & 0 & 1 & 0 & 1 & 0\\
    l & 0 & -l & 0 & 0 & -h\\
    -d_f & h & -d_f & h & d_r & 0\\
    0 & -l & 0 & l & 0 & -d_r
  \end{bmatrix}, \label{wrench matrix}
\end{equation}
where \(l\), \(h\), \(d_f\), \(d_r\) are the length of links described in \figref{trirotor dynamics}.
Each rotor has vectoring freedom, so the thrust force can be divided into horizontal (\(\lambda_{i, \parallel}\)) and vertical (\(\lambda_{i, \perp}\)) components.
\(\left|\bm{Q}\right| = -4l^2(d_f + d_r)\); hence, this cannot be \(0\) in this configuration. Therefore,
\(\bm{Q}\) is full rank and there exists an inverse matrix, and the thrust vector is calculated as follows
\begin{equation}
    \bm{\lambda} = \bm{Q}^{-1}
  \begin{bmatrix}
    \bm{f}\\\bm{\tau}
  \end{bmatrix}.
\end{equation}
The thrust and vectoring angle of each rotor can be calculated as follows
\begin{align}
  \lambda_i &= \sqrt{\lambda_{i, \perp}^2 + \lambda_{i, \parallel}^2},\\
  \alpha_i &= \tangent^{-1} \qty(\dfrac{\lambda_{i, \parallel}}{\lambda_{i, \perp}})\label{alpha}.
\end{align}

\subsection{Flying Humanoid with Passive Wheel on Foot}
\begin{figure}
  \centering
  \includegraphics[height=50mm]{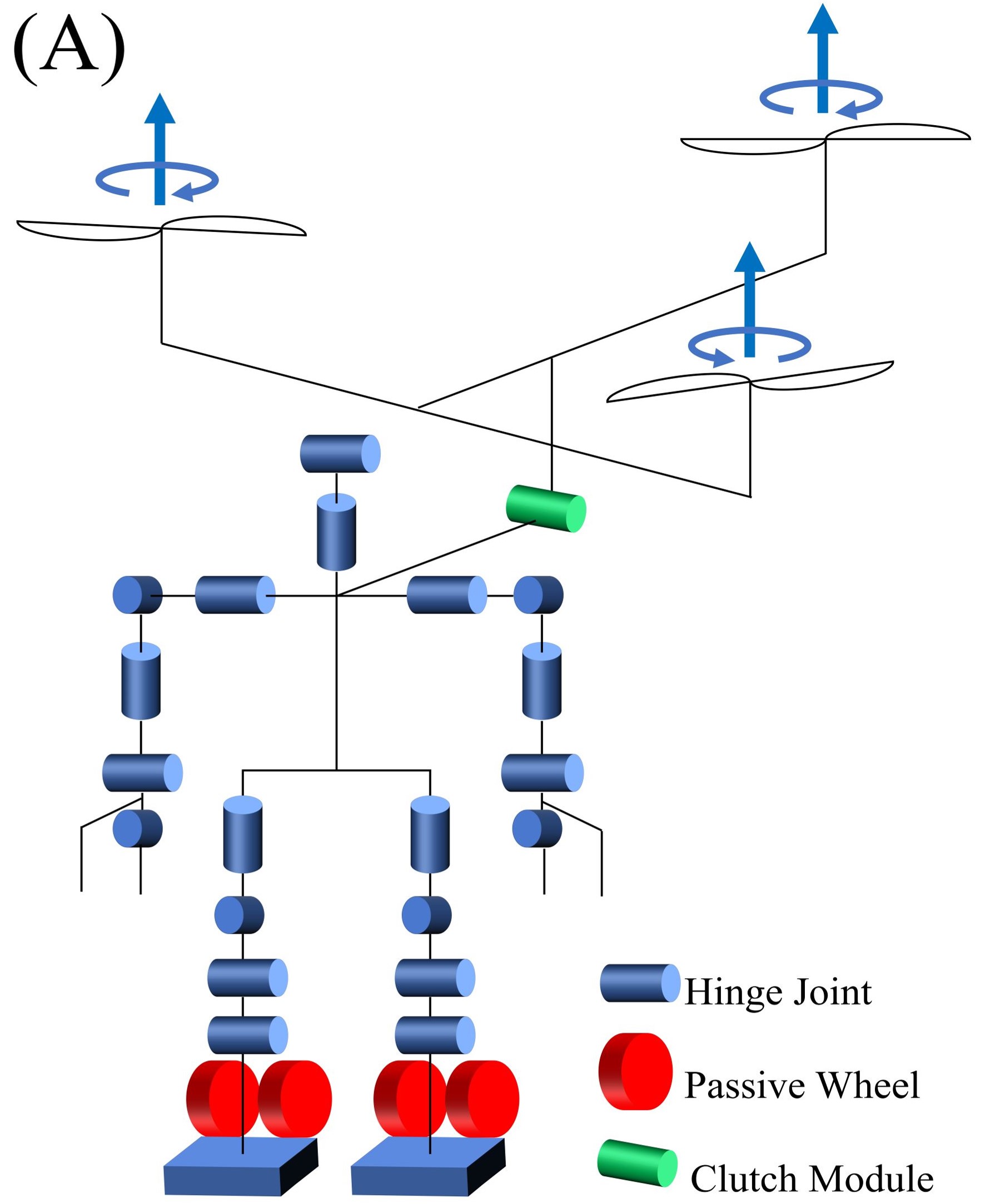}
  \includegraphics[height=50mm]{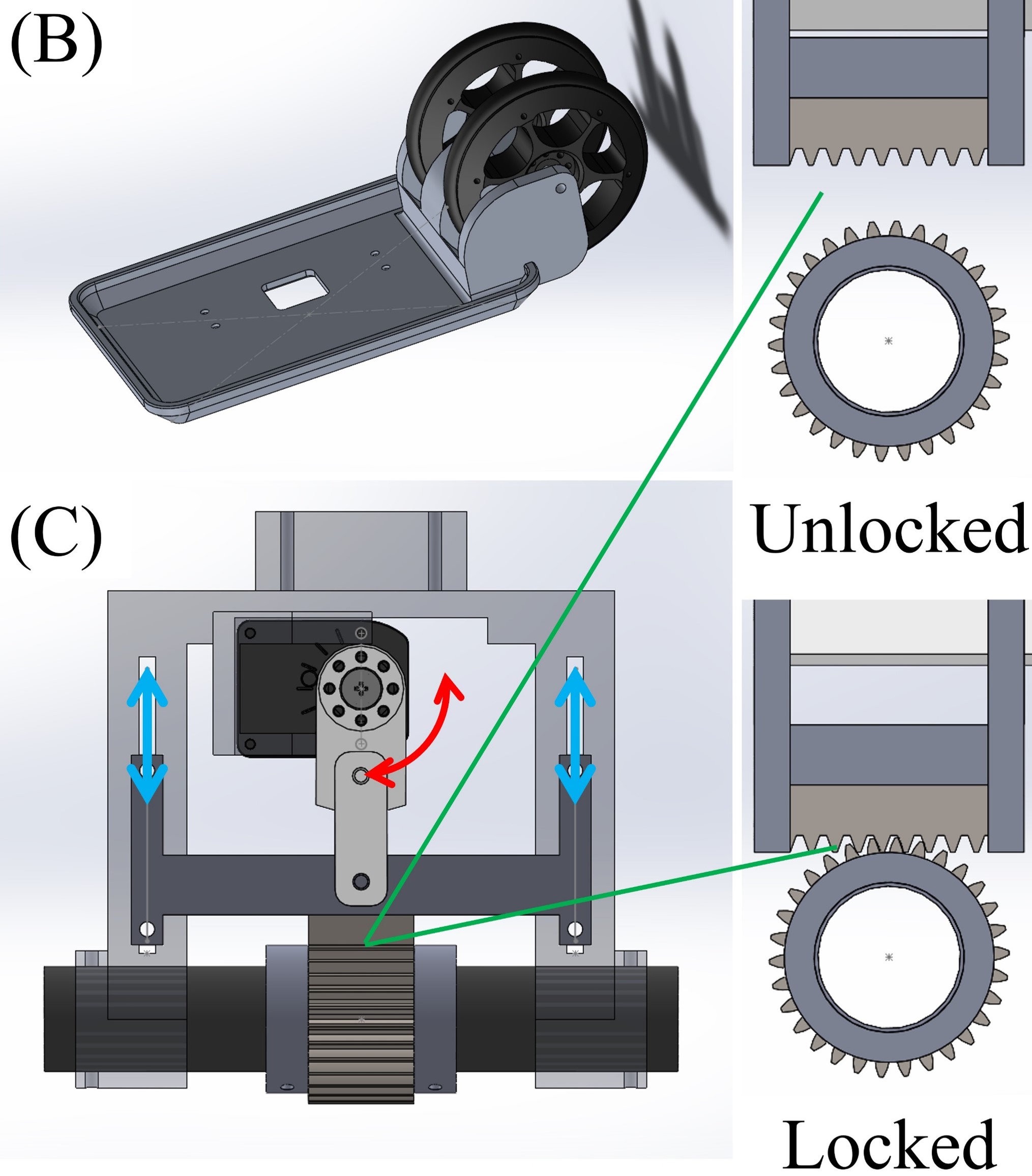}
  \caption{\textbf{Joint configuration and modules of flying humanoid}. (A): Joint configuration and arrangement of each module. Flight unit is deployed above the humanoid and clutch module is employed to achieve reconfigurable connection. Passive wheels are deployed on foot. (B): Foot with passive wheels. (C): Clutch module used to combine humanoid and flight unit.}
  \label{figure:humanoid configuration}
  \vspace{-3mm}
\end{figure}

In this work, the proposed humanoid is equipped with a flight unit to achieve terrestrial and aerial locomotion.
To avoid aerodynamic interference from the rotors on its arms during aerial manipulation, the flight unit is deployed behind the arms.
The joint configuration of the humanoid in this work is shown in \figref{humanoid configuration} (A).
Passive wheels are deployed on the foot to realize terrestrial locomotion, which is faster than walking by legs.
The passive wheels require minimal additional links and can be implemented without increasing the number of actuators, which is advantageous for aerial robots with payload constraints.
The foot design is shown in \figref{humanoid configuration} (B).
Wheels are deployed on the heel to avoid interference with arms in manipulation tasks such as picking up an object on the ground.
The distance between the foot plane and the radius of the wheels is the same; hence, by altering the other joints of the leg, the robot can switch between a state in which the foot plane is fully contacted and a state in which it is contacted only by the wheels.
When moving by wheels, maneuverability is achieved by thrust, thereby enabling the robot to move faster than walking with less energy consumption than flying.


\subsection{Clutch Module}\label{section:clutch}
In the connection between a humanoid and a flight unit, the relative pitch angle can be considered as a parameter.
In this work, we achieve a rotational DOF around the pitch axis in a connection.
In addition, the rotational DOF should be fixed to stabilize the motion of humanoid and aerial locomotion.
We designed a clutch module as shown in \figref{humanoid configuration} (C) for connection.
It can switch between a state in which it can rotate freely around a cylindrical axis and a state in which its rotation is fixed.
A crank mechanism driven by a servo motor is utilized to switch the rotational DOF by pressing a rack against a gear fixed around the rotational axis.
No torque is generated around the rotational axis of the actuator to fix; hence, this mechanism is durable.
This module is deployed on the back of the humanoid, as shown in \figref{humanoid configuration} (A).
The relative pitch angle is determined in each mode of locomotion based on the optimization method in \secref{opt} to achieve a robust state.

\subsection{Optimal Orientation of Flight Unit}
\label{section:opt}
\begin{figure}[t]
  \centering
  \includegraphics[width=0.6\columnwidth]{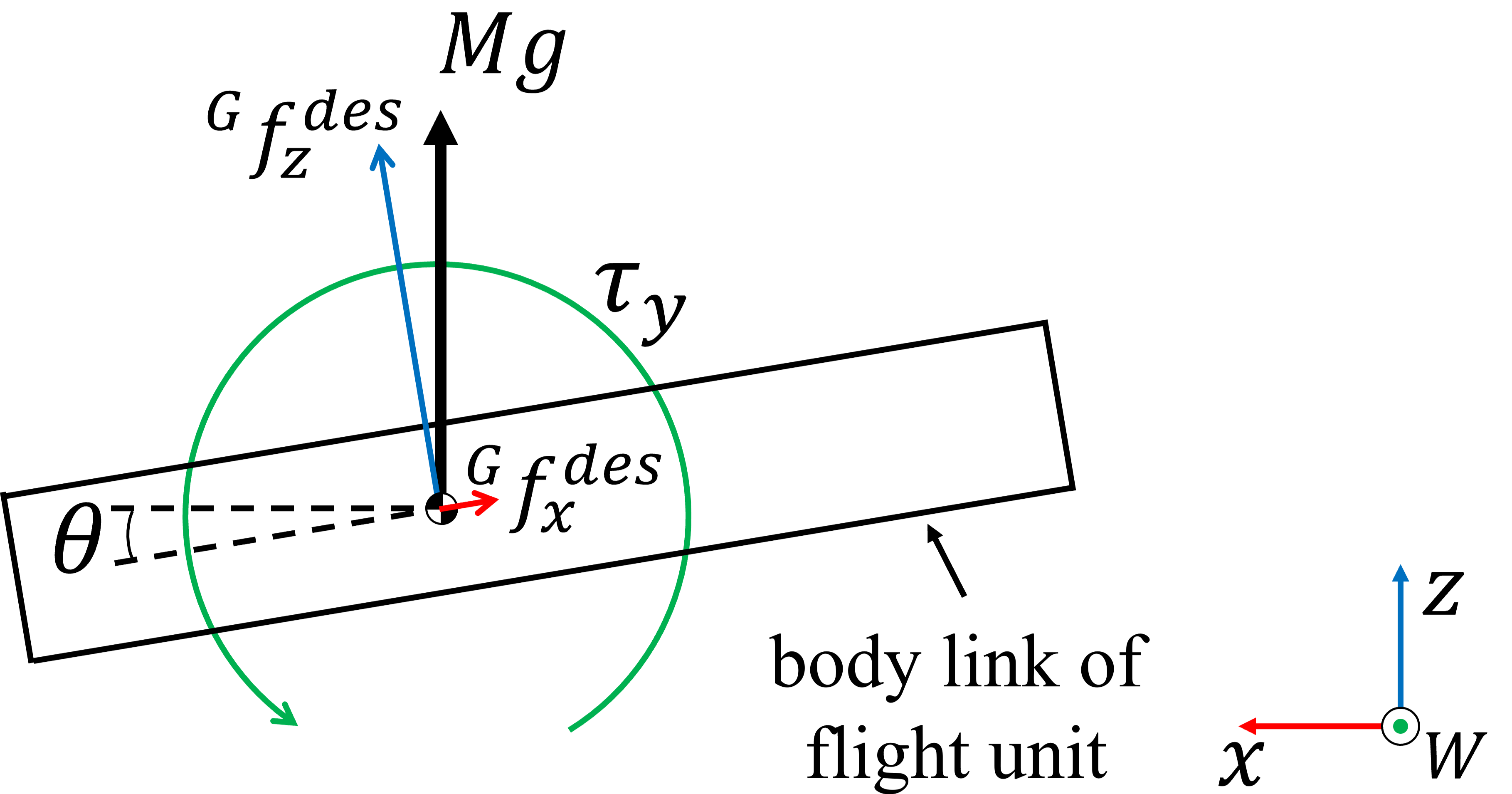}
  \vspace{-3mm}
  \caption{\textbf{Desired wrench described in \eqref{optimization wrench}}. Gravity compensation term is divided into x and z directions, and torque \(\tau_y\) is exerted.}
  \label{figure:wrench}
  \vspace{-3mm}
\end{figure}

\begin{figure}
  \centering
  \includegraphics[width=0.9\columnwidth]{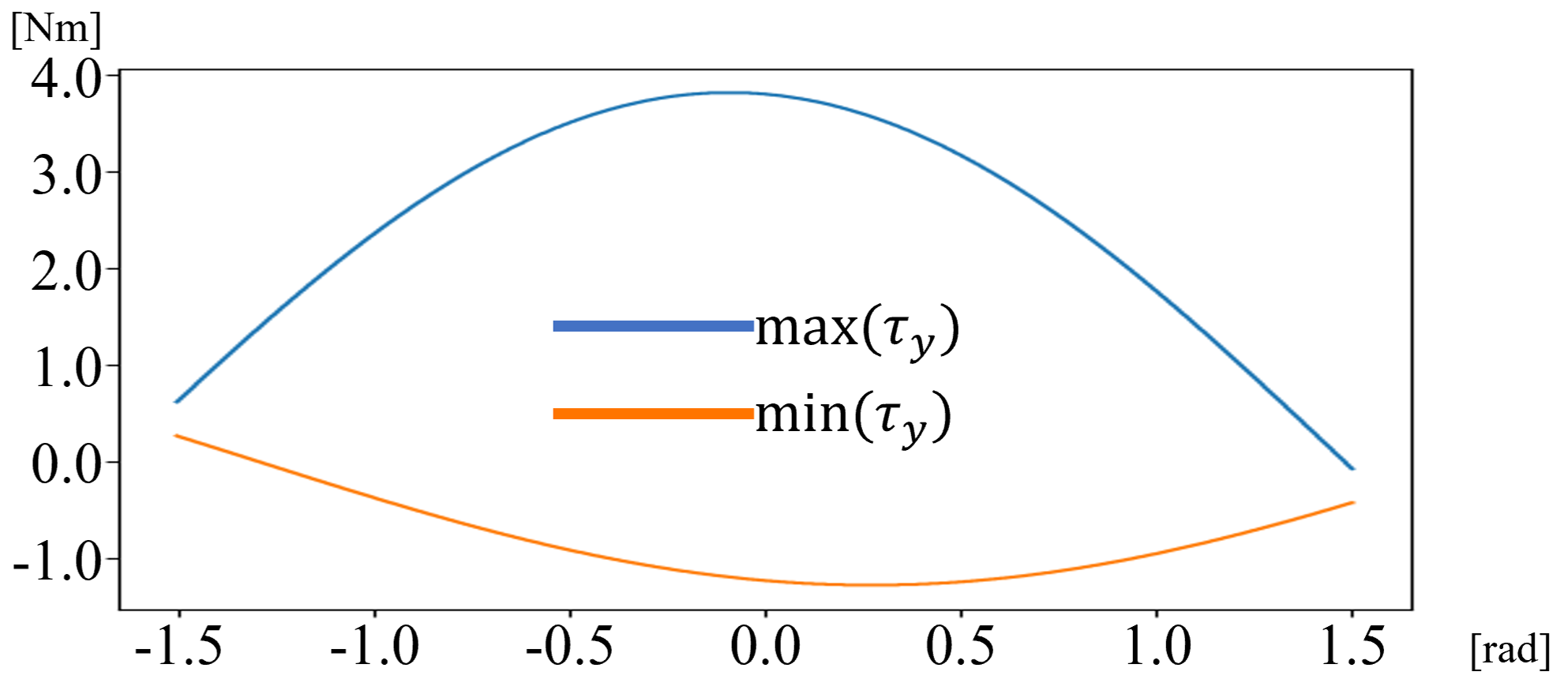} 
  \vspace{-3mm}  
  \caption{\textbf{Plot of feasible torque around pitch axis when tilting by pitch angle}. Blue line: \(\text{max}\qty(\tau_y)\) in \eqref{opt}. Orange line: \(\text{min}\qty(\tau_y)\) in \eqref{opt}. The range of feasible torque is maximized when the pitch angle of the body is 0 \si{rad}}.
  \label{figure:trirotor feasible torque}
  \vspace{0mm}
\end{figure}

Using the clutch module described in \secref{clutch}, the relative angle in the connection between the humanoid and flight unit becomes reconfigurable.
We demonstrate the optimal attitude of the body link of the trirotor in terms of the robustness of the control.
In a previous work, the arrangement of rotors was optimized to maximize feasible force and torque \cite{Park2018odar}.
In this work, we only focus on pitch rotation and design an optimization problem that obtains the optimal pitch angle to maximize the range of the feasible torque because the structure is symmetrical around other axes.
The optimization problem can be obtained as follows

\begin{align}
  &\underset{\theta}{\text{maximize} }&&\max\qty(\tau_y) - \min\qty(\tau_y),\label{opt}\\
  &\text{subject to} && 
 \bm{w}^{des}= \bm{Q} \bm{\lambda}, \\ 
  &&&
  \bm{w}^{des} =
  \begin{bmatrix}
      -Mg\sin\theta\\0\\Mg\cos\theta\\0\\\tau_y\\0
  \end{bmatrix}, \label{optimization wrench}\\    
  &&&\lambda_i = \sqrt{\lambda_{i, \perp}^2 + \lambda_{i, \parallel}^2},\\
  &&&\alpha_i = \tangent^{-1}\qty(\frac{\lambda_{i, \parallel}}{\lambda_{i, \perp}}),\\
  &&&\lambda_{min} < \lambda_i < \lambda_{max},\label{lambda constraints}\\
  &&&\alpha_{min} < \alpha_i < \alpha_{max}.\label{alpha constraints}
\end{align}  
\eqref{optimization wrench} suggests that when the flight unit is tilted by a \(\theta\) pitch angle, the force required for hovering is divided into the x and z axes in the CoG frame, as shown in \figref{wrench}.
Besides, the target torque around pitch axis is \(\tau_y\).
Furthermore, \eqref{lambda constraints} and \eqref{alpha constraints} are constraints on the upper and lower limits of the thrust and vectoring angle, thereby setting thresholds that are feasible for real actuators.
Feasible \(\tau_y\) is calculated for each \(\theta\), and \(\text{max}\qty(\tau_y)\) and \(\text{min}\qty(\tau_y)\) in \eqref{opt} are their maximum and a minimum values, respectively and shown in \figref{trirotor feasible torque}.
The \(\theta_{flight\_unit}\) that maximizes the range of feasible torque around the pitch axis is 0 \si{rad}. 
In this state, the control margin is maximized, such that the most robust state is achieved. 
In this work, we solved this optimization problem by examining all possible actuator inputs because the scale of the problem is small and need not to be solved online during realtime control.

\subsection{Desired Clutch Angle}
\label{section:pose}
\begin{figure}
  \centering
  \includegraphics[width=0.4\columnwidth]{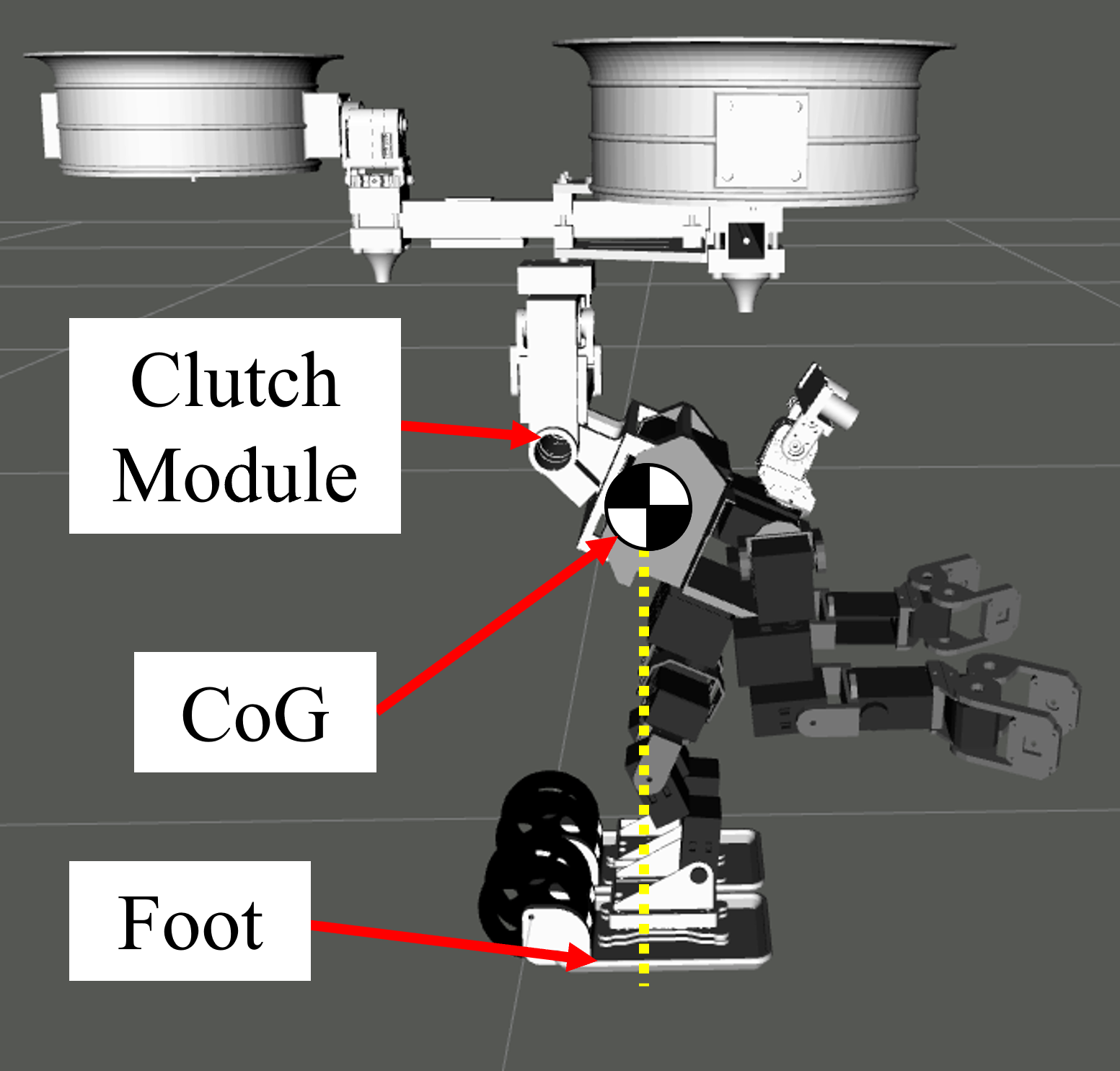}
  \includegraphics[width=0.4\columnwidth]{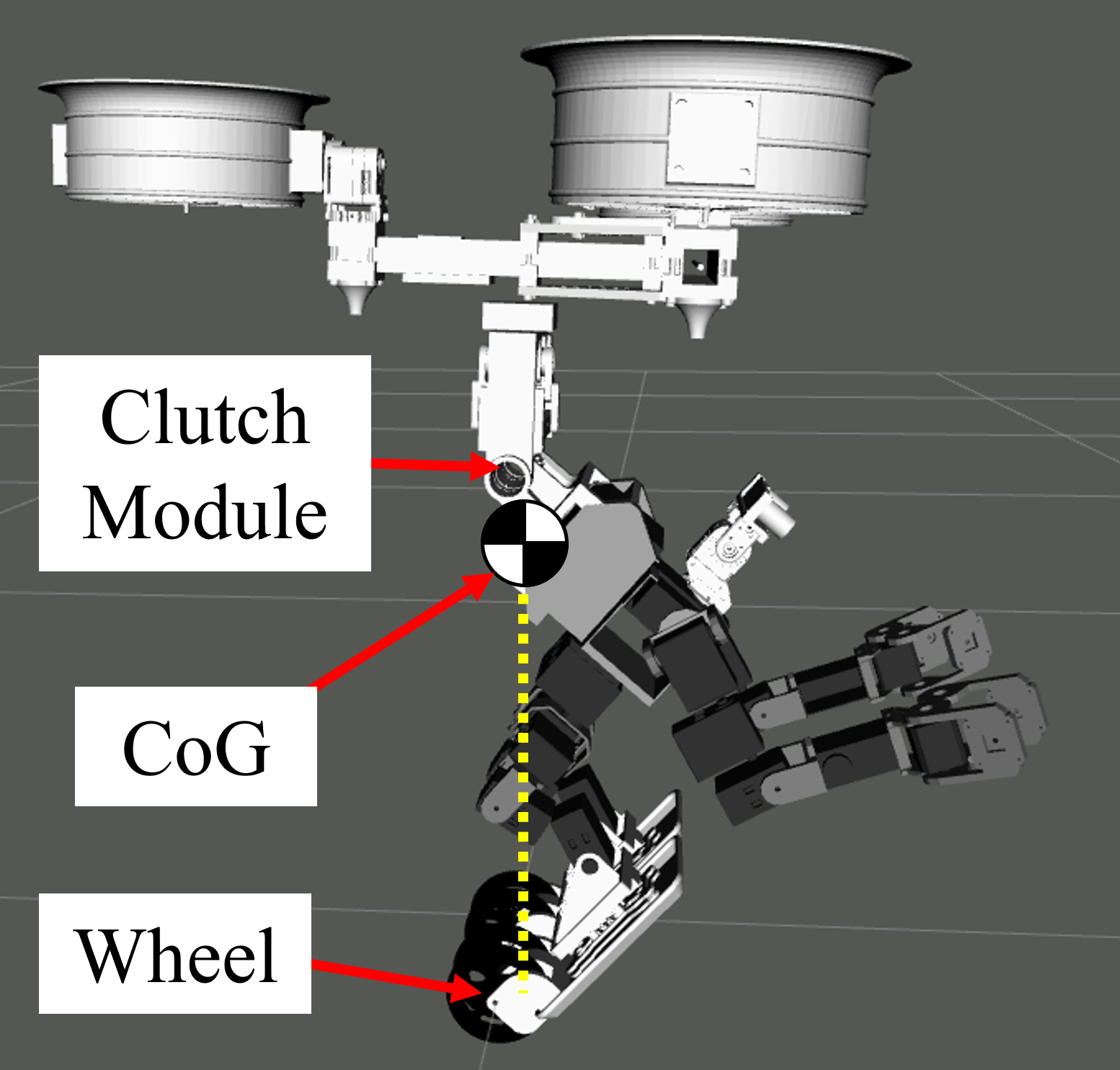}
  \begin{minipage}{0.4\hsize}\centering (A)\end{minipage}
  \begin{minipage}{0.4\hsize}\centering (B)\end{minipage}
  \vspace{-3mm}
  \caption{\textbf{Feasible pose in each locomotion mode}. (A): Leg locomotion. Support polygon area is rectangle made by foot plate. (B): Wheel locomotion. Support polygon area is a line segment connecting ground contact points.}
  \label{figure:desired pose}
  \vspace{-4mm}
\end{figure}

In \secref{opt}, we present the optimal orientation of the flight unit.
The joint angles of the humanoid are changed depending on the mode of locomotion.
It is required to determine the appropriate clutch angle under each mode of locomotion based on the position of the CoG and foot contact condition.
In each mode of locomotion, \(\theta_{clutch}\), which is the relative pitch angle between the humanoid and flight unit, should be in the following set
\begin{equation}
  \label{desire angle}
  \theta_{clutch}\! \in \! \qty{\theta_{torso}\qty(\bm{q}) - \theta_{flight\_unit} \mid CoG\qty(\bm{q}) \in S \qty(\bm{q})},
\end{equation}
where \(\bm{q}\), \(\theta_{flight\_unit}\), \(\theta_{torso}\), and \(CoG\) denote the joint angle vector of the humanoid, optimal pitch angle of the flight unit described in \secref{opt}, pitch angle of torso link of humanoid, and orthographic projection of the humanoid's CoG to the ground, respectively.
\(\theta_{torso}\) and \(CoG\) are calculated by forward kinematics using a robot model and \(\bm{q}\).
\(S\) represents the support polygon area of each mode of locomotion and is described as follows.
In the leg locomotion mode, \(S\) is a support rectangle of feet.
In the wheel locomotion mode, \(S\) is a line segment connecting the ground contact points.
One of the valid poses in leg locomotion and wheel locomotion are depicted in \figref{desired pose} (A) and \figref{desired pose} (B).
In the aerial locomotion mode, the valid pose is the same as the leg locomotion mode because the foot should be horizontal to the ground for stable takeoff and landing.

\section{CONTROL}
\label{section:3}
\begin{figure}
  \centering
  \includegraphics[width=1.0\columnwidth]{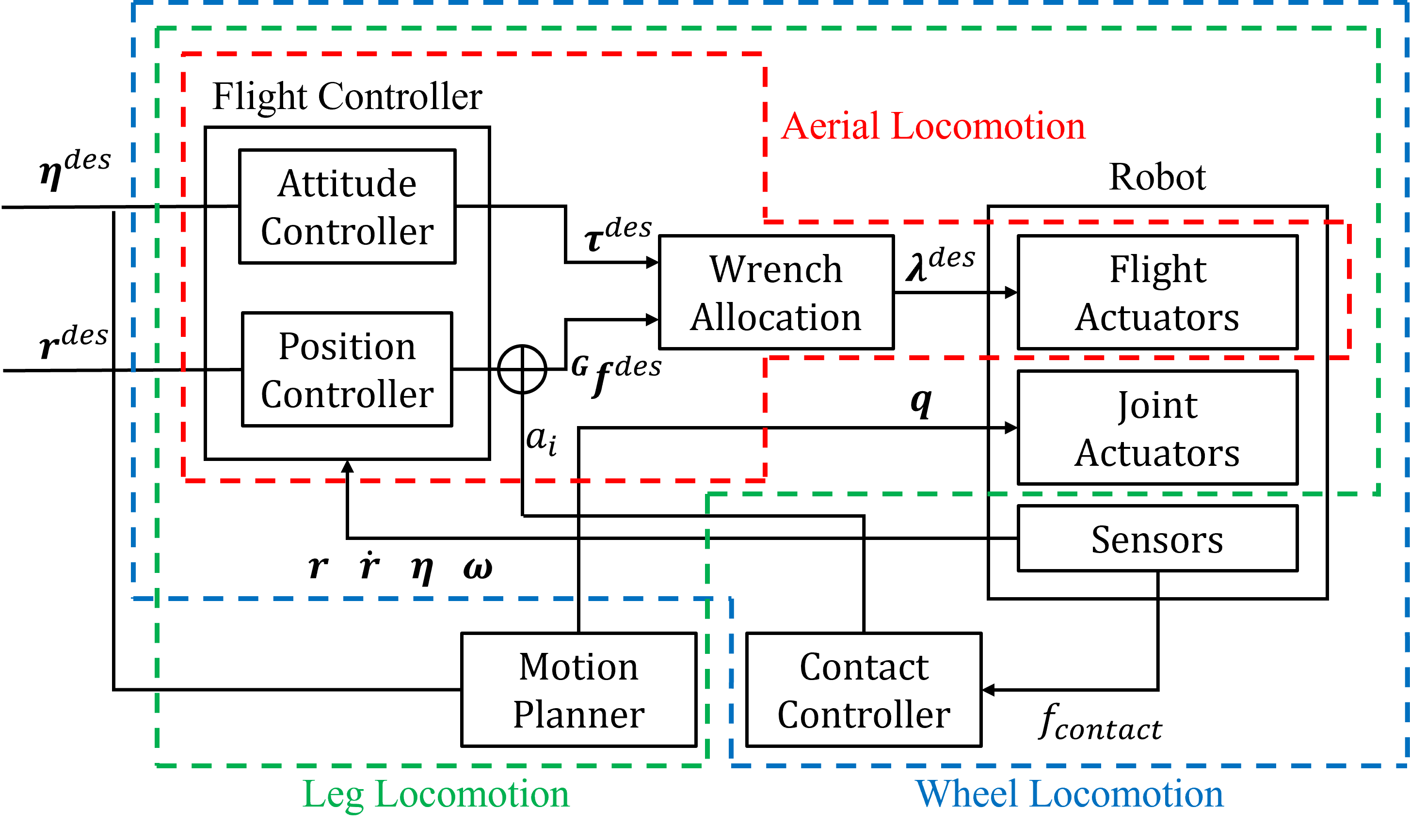}
  \vspace{-8mm}
  \caption{\textbf{Integrated control framework of flying humanoid}. Legged and wheeled locomotions are  based on flight control for aerial locomotion.}
  \label{figure:flight control}
  \vspace{-3mm}
\end{figure}

In this section, an integrated control framework shown in \figref{flight control} is presented for the proposed flying humanoid.
First, we present flight control comprising attitude control, position control, and wrench allocation.
Flight control is also employed in legged and wheeled locomotions.
Then, the motion planning method for leg locomotion and the contact control method for wheel locomotion are described.

\subsection{Aerial Locomotion}
Flight control comprises attitude control, position control, and wrench allocation.

\subsubsection{Attitude Control}
The target torque in the {CoG} frame can be calculated as follows.
\begin{align}
  \bm {\tau }^{des} = &\bm{\omega}_{\lbrace CoG \rbrace} \times \bm{I} \bm{\omega}_{\lbrace CoG \rbrace} + \bm {K}_{\eta, p} \bm{\eta}^{err} \nonumber\\
  &+\bm{K}_{\eta, i} \int \bm{\eta}^{err} \mathrm{d}t + \bm{K}_{\eta, d} \bm{\omega}_{\lbrace CoG \rbrace}^{err},
\end{align}
\begin{equation}
  \bm {\eta}^{err} = \bm {\eta}^{des} - \bm {\eta},
\end{equation}
where \(\bm{\eta}\), $\bm {K}_{\eta, *} \in \mathbb{R}^{3 \times 3}$ denote the euler angle vector of the CoG frame in the world frame and PID gain diagonal matrices, respectively.

\subsubsection{Position Control}
The target translational force in the CoG frame can be calculated as follows
\begin{align}
  \bm{f}^{des} = &M \bm{R}^{-1} \bigg( \bm{g} + \bm{K}_{r, p} \bm{r}^{err} + \bm{K}_{r, i}\int \bm{r}^{err} \mathrm{d}t \color{white}\nonumber\\
  &+ \bm{K}_{r, d} \dot{\bm{r}}^{err} \bigg),\label{chap03: position control}
\end{align}
\begin{equation}
  \bm{r}^{err} = \bm{r} ^{des} - \bm{r},
\end{equation}
where \(\bm{K}_{r, *} \in \mathbb{R}^{3\times3}\) denote the PID gain diagonal matrices.

\subsubsection{Wrench Allocation}
Target thrust and the vectoring angle of each rotor are calculated from the target torque and translational force in the CoG frame using \eqref{wrench allocation}--\eqref{alpha}.

\subsection{Leg Locomotion}
\label{section:leg locomotion}
We utilized the thrust exerted by flight control to stabilize the walking motion.
In this work, walking motion is generated using inverse kinematics beforehand and joints follow its trajectory.
In each footstep motion during walking, desired roll and yaw angles of flight control are updated as follows
\begin{align}
  \phi^{des} &\leftarrow \phi_{torso}\qty(\bm{q}),\\
  \psi^{des} &\leftarrow \psi_{torso}\qty(\bm{q}),
\end{align}
where \(\phi_{torso}\qty(\bm{q})\) and \(\psi_{torso}\qty(\bm{q})\) denote the roll and yaw angles of the torso link of the humanoid when its support leg is parallel to the ground.
They are calculated by forward kinematics using a robot model and joint angle vector \(\bm{q}\).
Besides, the target x-y position is updated to the position at the time.
The target position of z is set above the position of the flight unit to avoid falling.
Furthermore, linear acceleration in the z-direction should be larger than \(\alpha_{stable}\) to decrease load torque on leg joints and raise the swing leg stably.
\(\alpha_{stable}\)  was determined experimentally for stable motion.
In addition to this constraint, thrust should be sufficiently small to avoid hovering; hence, the following clamp is added in the position control

\begin{equation}
  \label{leg clamp}
  \alpha_{stable} < \ddot{r}_{z} < g.
\end{equation}

\subsection{Wheel Locomotion}
\label{section:wheel locomotion}
In the wheel locomotion mode, the foot and ground make point contact and maneuvering is more difficult than in leg locomotion.
We introduce feedback control to maintain stable contact with the ground during the wheeled locomotion.
Contact force with the ground is measured by force sensors installed on the foot.
A feedback term \(a_i\) is introduced in position control of z.
The position control while wheel locomotion is described as follows
\begin{equation}
  \ddot{r}_{z} = k_{p, z} z^{err} + k_{i, z} \int z^{err} + k_{d, z} z^{err} + a_i,
\end{equation}
where \(k_{p, *}\) are PID control gains.
\(a_i\) is updated in the control loop as follows
\begin{equation}
  a_{i+1} = a_{i} + k_f \frac{f_{contact} - f_{thresh}}{M},
\end{equation}
where \(k_f\), \(f_{contact}\), and \(f_{thresh}\) denote feedback gain, contact force, and target contact force, respectively.

In addition to this feedback term, the following clamp is added because it is necessary to exert the minimum thrust to prevent falling and hovering just like \eqref{leg clamp}
\begin{equation}
  \label{wheel clamp}
  \beta_{stable} < \ddot{r}_z < g,
\end{equation}
where \(\beta_{stable}\) denotes the minimum linear acceleration required to avoid falling, obtained experimentally as \(\alpha_{stable}\) in \eqref{leg clamp}.

\section{EXPERIMENT}
\label{section:experiment}
In this section, we present a prototype of the flying humanoid robot proposed in this work.
Then, we conduct locomotion experiments in each mode: aerial, legged, and wheeled locomotions.
Finally, object transportation and aerial manipulation experiments are conducted.

\begin{figure}[t]
  \centering
  \includegraphics[width=1.0\columnwidth]{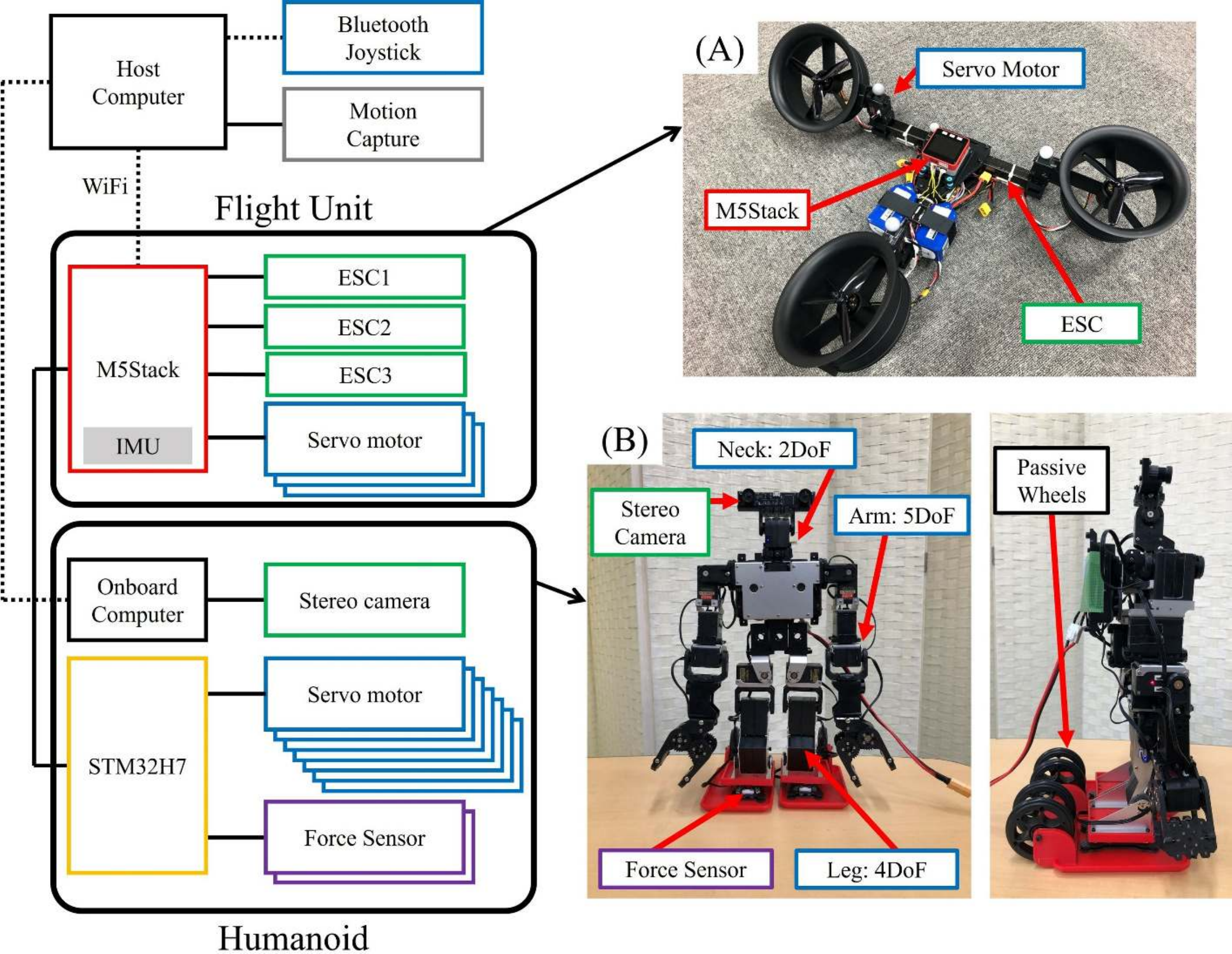}
  \vspace{-7mm}
  \caption{\textbf{Hardware and system}. Left: Diagram of hardware system. (A): Fully-actuated trirotor flight unit. It has three EDFs and three servo motors. M5Stack equipped with IMU works as a flight controller. (B): KHR-W. It has 20 DoFs and force sensor in each foot.}
  \label{figure:hardware system}
  \vspace{-4mm}
\end{figure}

\subsection{Robot Platform}
The total hardware system is shown in \figref{hardware system}.
The prototype of the proposed flying humanoid comprises a fully-actuated trirotor flight unit (\figref{hardware system} (A)) and a humanoid robot KHR-W (\figref{hardware system} (B)).

\subsubsection{Flight Unit}
The fully-actuated trirotor flight unit has three electric ducted fans (EDFs) and three electric speed controllers (ESCs).
In this work, we developed a trirotor flight unit shown in \figref{hardware system} (A), and the specifications are presented in \tabref{trirotor spec}.
The maximum thrust of the EDF is 18 \si{N}.
Servo motors for vectoring control of rotors were KRS-3304R2 ICS.
The body of the trirotor is made of a CFRP square pipe for a lightweight design.
The rest of the body is made of PLA.
M5Stack with an ESP32 microcontroller (2 cores, 240 \si{MHz}) was deployed to perform the realtime pose control (100 \si{Hz}) as presented in \figref{flight control}.
IMU in M5Stack was employed to estimate \(\bm{\eta}\) and \(\bm{\omega}_{\lbrace CoG \rbrace}\) for this realtime control.
In addition, the external motion capture system (sampling rate: 100\si{Hz}), with an error margin less than a few millimeters was applied to obtain \(\bm{r}\) and \(\dot{\bm{r}}\).
For evaluation, \(\bm{\eta}\) from the motion capture data was also used.




\begin{table}[t]
  \centering
  \caption{Trirotor Flight Unit}
  \begin{tabular}{c|c}
    Attribute & Value\\\hline
    rotor KV & \num{1550}\\
    max rotor thrust & \num{18} \si{N}\\
    max vectoring torque & \num{1.36} \si{Nm}\\
    vectoring angle range & [-3/4\(\pi\), 3/4\(\pi\)] \si{rad}\\
    mass & \num{1652} \si{g}\\
    power & \num{22.2} \si{V}\\
    size & \num{640} \(\times\) \num{463} \(\times\) \num{124}\si{mm}
  \end{tabular}
  \label{table:trirotor spec}
  \vspace{-2mm}
\end{table}

\begin{table}[t]
  \centering
  \caption{KHR-W}
  \begin{tabular}{c|c}
    Attribute & Value\\\hline
    leg joint max torque & \num{2.45} \si{Nm}\\
    arm joint max torque & \num{1.37} \si{Nm}\\
    power & \num{12} \si{V}\\
    mass & \num{1691} \si{g}\\
    size & \num{170} \(\times\) \num{160} \(\times\) \num{350} \si{mm}
  \end{tabular}
  \label{table:khr spec}
  \vspace{-3mm}
\end{table}

\subsubsection{Humanoid KHR-W}
In this work, we developed a humanoid robot KHR-W as shown in \figref{hardware system} (B).
\tabref{khr spec} presents the specifications of KHR-W.
This is made from KHR, a hobby robot product by Kondo Kagaku Co. Ltd.
KRS-2552R2HV ICS is used for 2 DoF neck and 5 DoF arms, and KRS-2572R2HV ICS is adopted for 4 DoF legs.
It also has two passive wheels in each foot as shown in \figref{humanoid configuration}.
Force sensors proposed in \cite{TransformHumanoid:Makabe:ICRA2022} are deployed on the foot for feedback control described in \secref{wheel locomotion}.
To control the joint servo motors and force sensors, the MCU board with STM32H7 (1 core, 480 \si{MHz}) is in the back of the body.
A stereo camera module on the head and UPBoard with the Intel Atom x5-Z8350 processor (4 cores, 1.92 \si{GHz}) at the back of the body are deployed to obtain an image of view and process it.
The clutch module shown in \figref{humanoid configuration} was at the back of the KHR-W. This module is fixed to an optimal angle.
When the locomotion mode is changed and the foot is contacted to the ground, it is reconfigurable by the attitude control of the flight unit.




\begin{figure}[t]
  \centering
  \includegraphics[width=1.0\columnwidth]{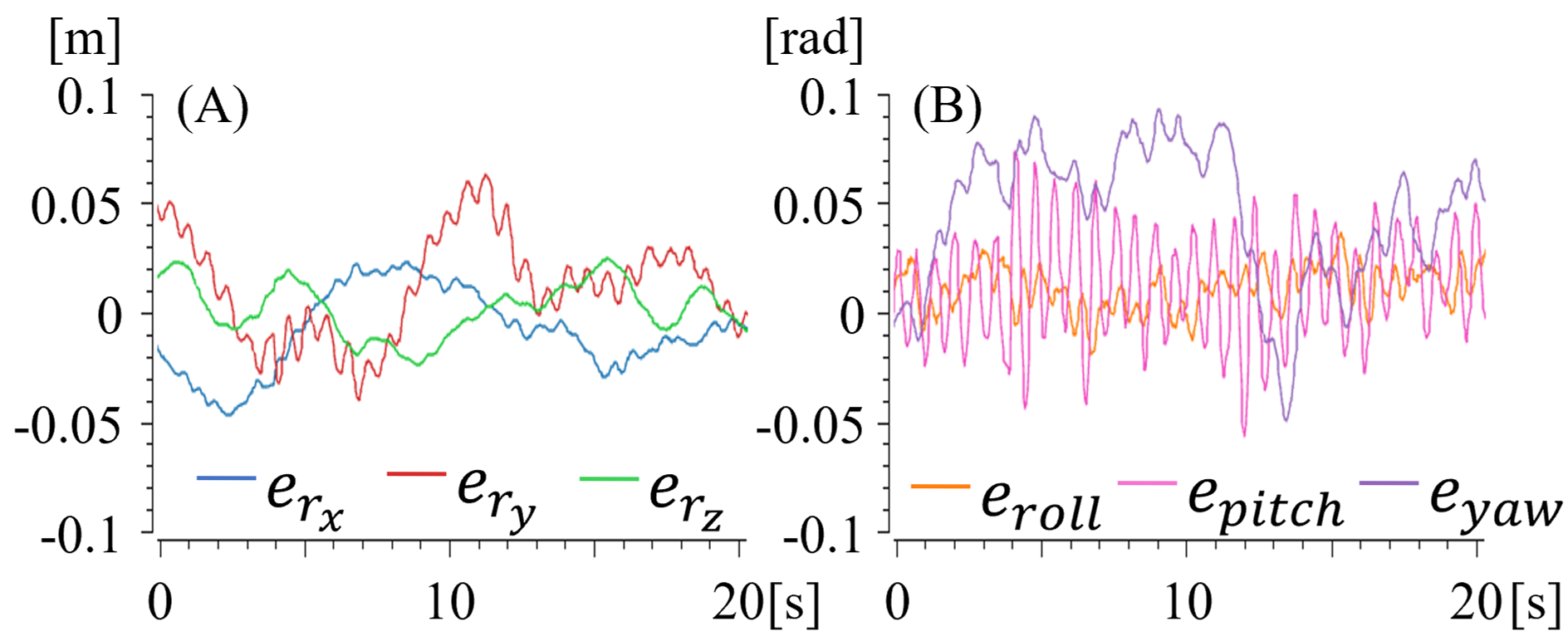}
  \vspace{-8mm}
  \caption{\textbf{Plots of hovering experiment}. (A): Errors of position control. (B): Errors of attitude control.}
  \label{figure:hovering plot}
  \vspace{-4mm}
\end{figure}

\begin{figure}[b]
  \centering
  \includegraphics[width=0.30\columnwidth]{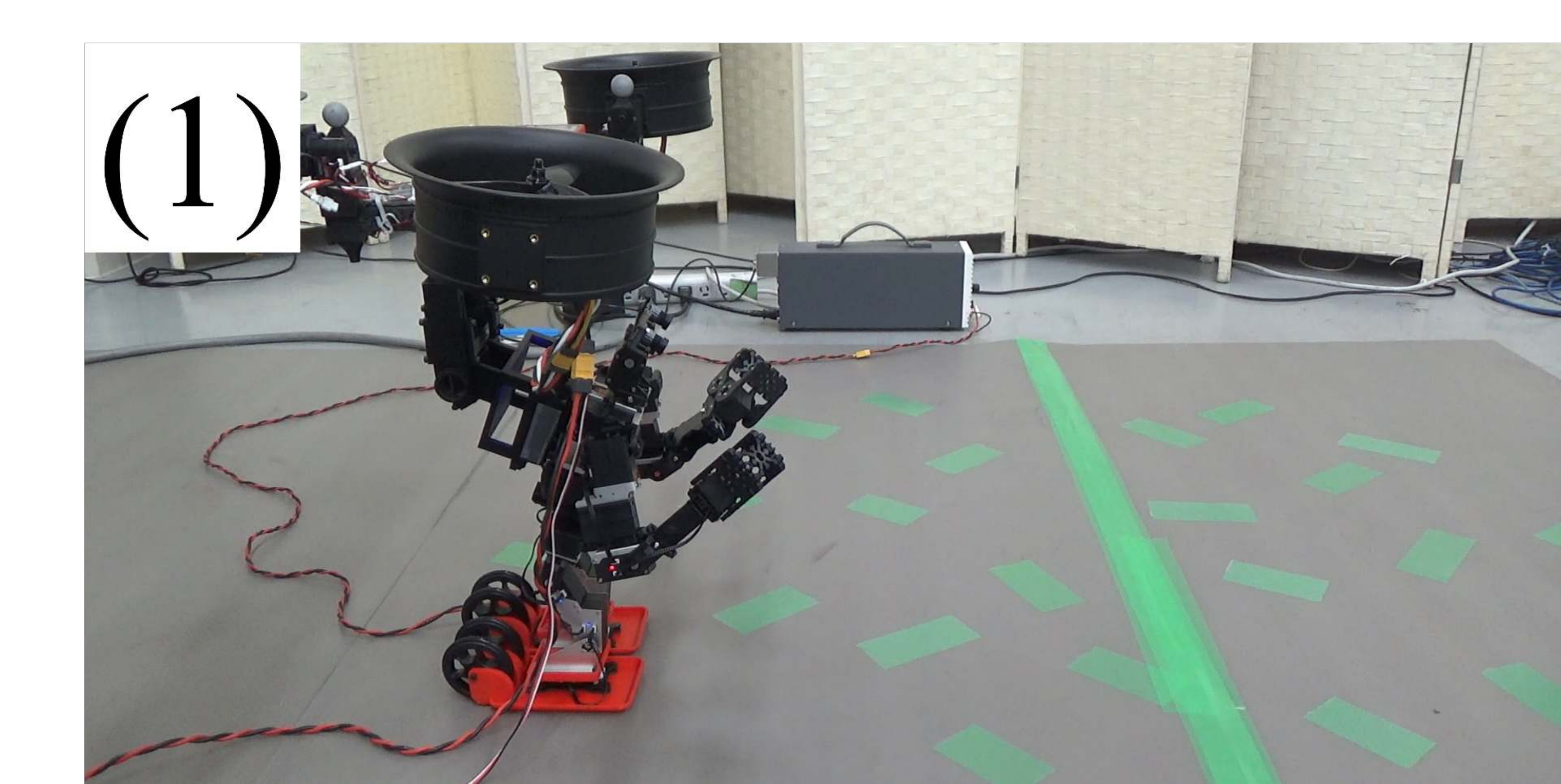}
  \includegraphics[width=0.30\columnwidth]{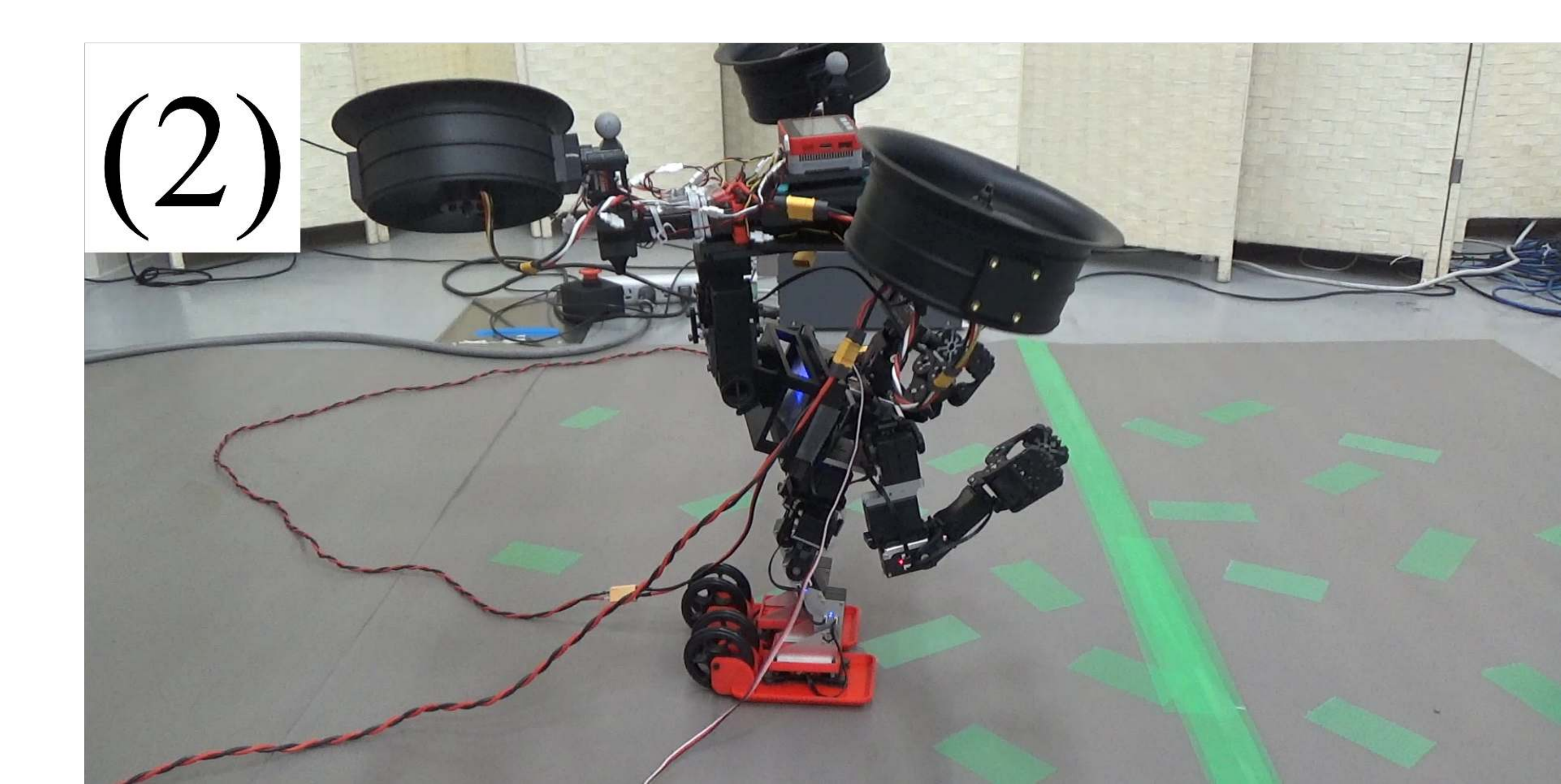}
  \includegraphics[width=0.30\columnwidth]{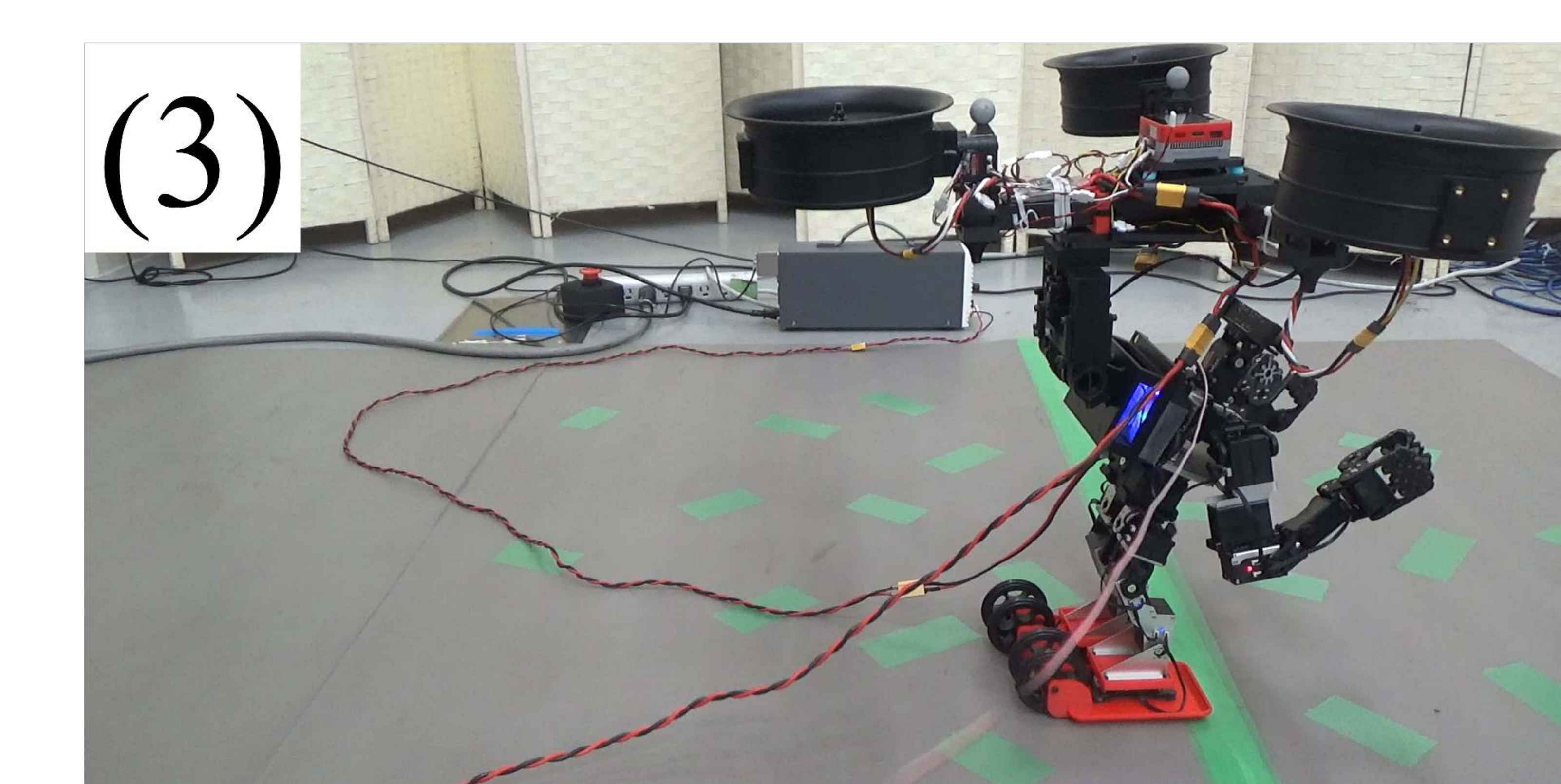}
  \vspace{-2mm}
  \caption{\textbf{Leg locomotion experiment}. Using thrust of flight unit, walking long distance without falling was achieved.}
  \label{figure:walking experiment}
  \vspace{-2mm}
\end{figure}

\begin{figure}[b]
  \centering
  \includegraphics[width=1.0\columnwidth]{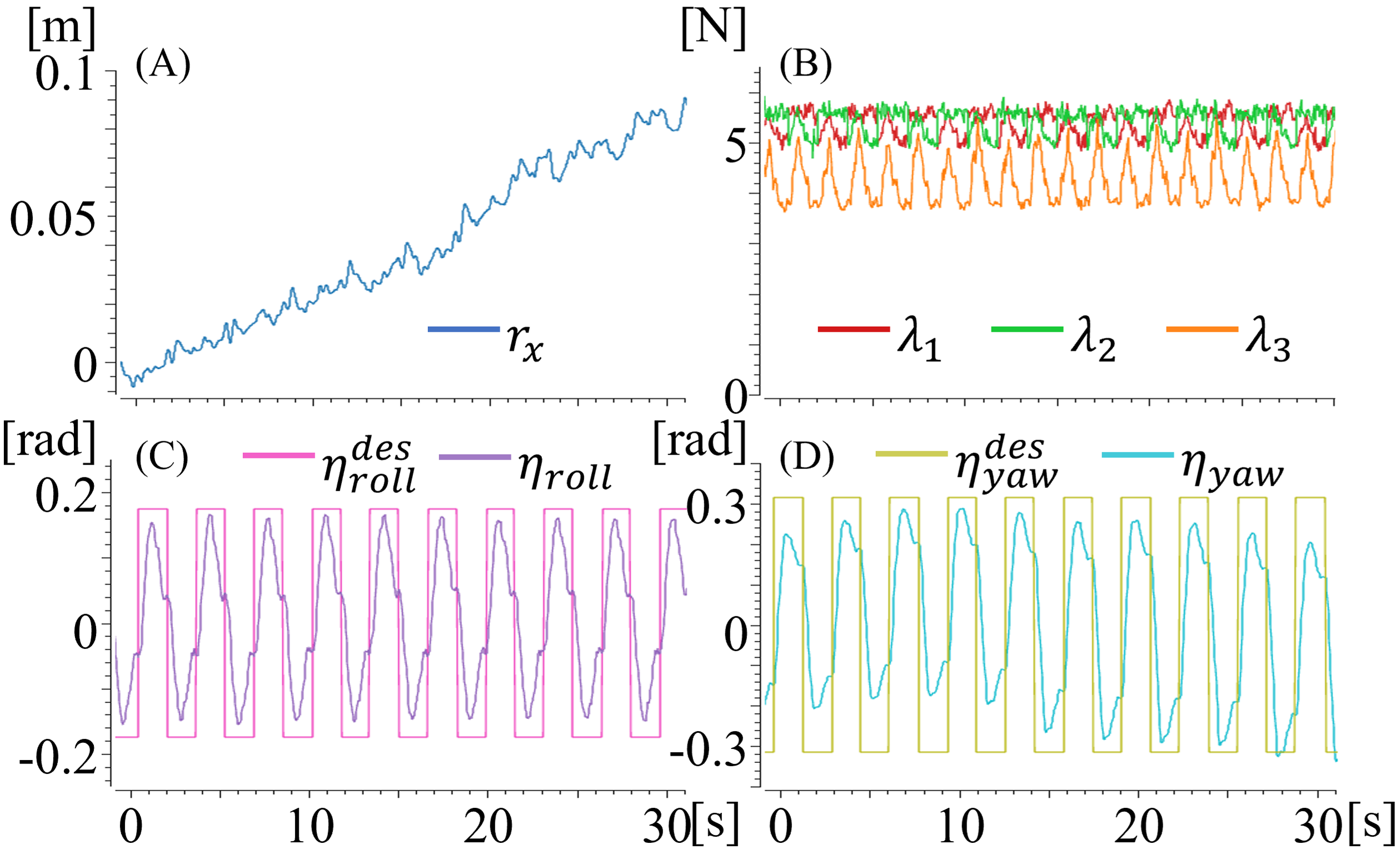}
  \vspace{-8mm}
  \caption{\textbf{Plots related to \figref{walking experiment}}. (A): Position x of flight unit. (B): Thrusts. (C): Target and actual roll angles of flight unit. (D): Target and actual yaw angles of flight unit.}
  \label{figure:walking plot}
  \vspace{-3mm}
\end{figure}

\subsection{Locomotion Experiment}
\subsubsection{Aerial locomotion}
A hovering experiment was conducted to evaluate the stability of aerial locomotion.
\(\theta_{clutch}\) that is the relative pitch angle between the humanoid and flight unit described in \secref{pose} was 25\degree{} in this experiment.
This angle was calculated by substituting \(\theta_{flight\_unit}\) and joint angle \(\bm{q}\) into \eqref{desire angle}.
\figref{hovering plot} presents the position and attitude errors during hovering.
The RMS of these errors were \(\begin{bmatrix}
  0.0208 & 0.0262 & 0.0125
\end{bmatrix} \si{m}\)
and 
\(\begin{bmatrix}
  0.0147 & 0.0278 & 0.0557
\end{bmatrix} \si{rad}\).
The maximum absolute values of these errors were \(\begin{bmatrix}
  0.0471 & 0.0401 & 0.0239
\end{bmatrix} \si{m}\)
and
\(\begin{bmatrix}
  0.0369 & 0.0735 & 0.0930
\end{bmatrix} \si{rad}\).


\begin{figure}[b]
  \centering
  \includegraphics[width=0.30\columnwidth]{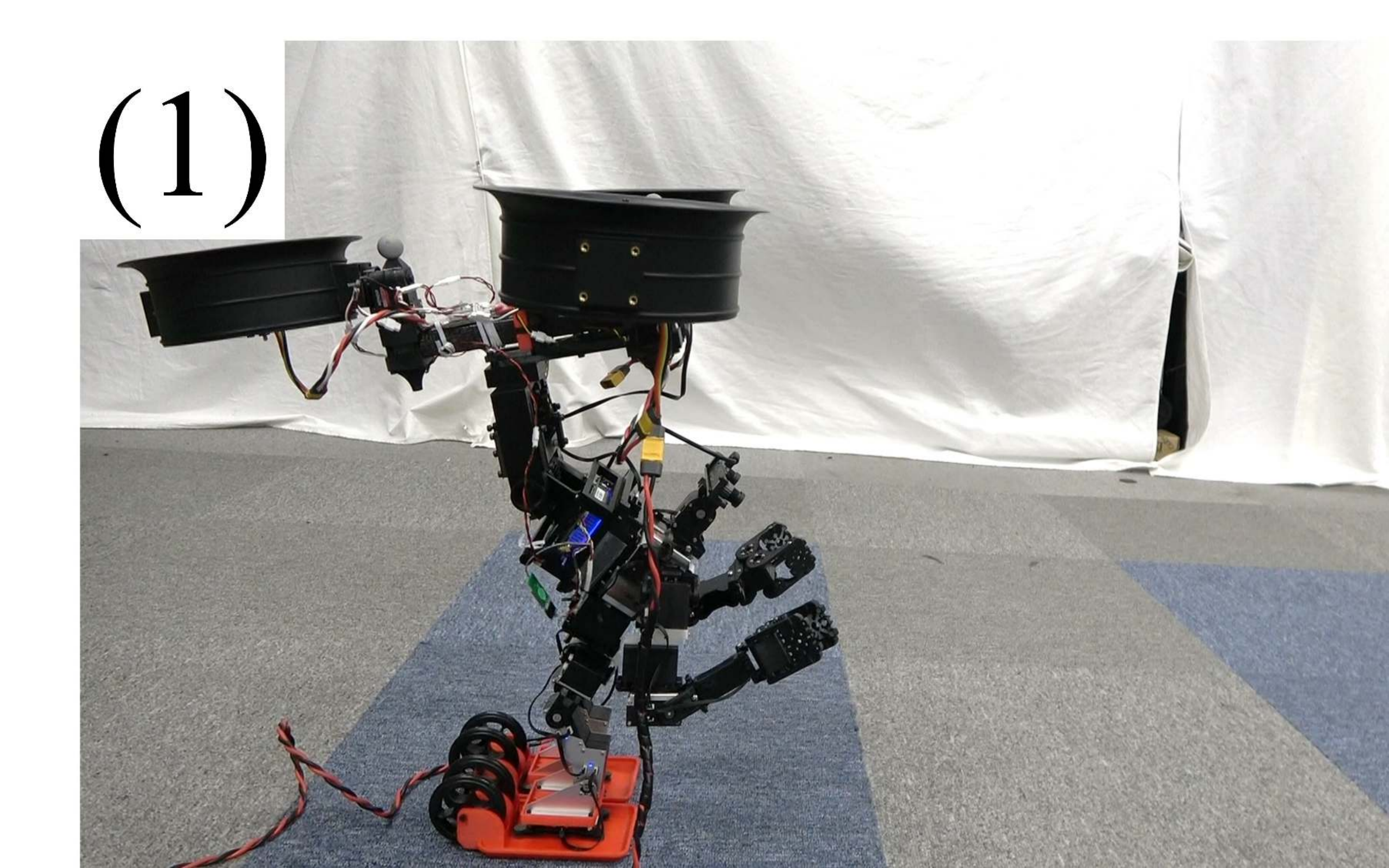}
  \includegraphics[width=0.30\columnwidth]{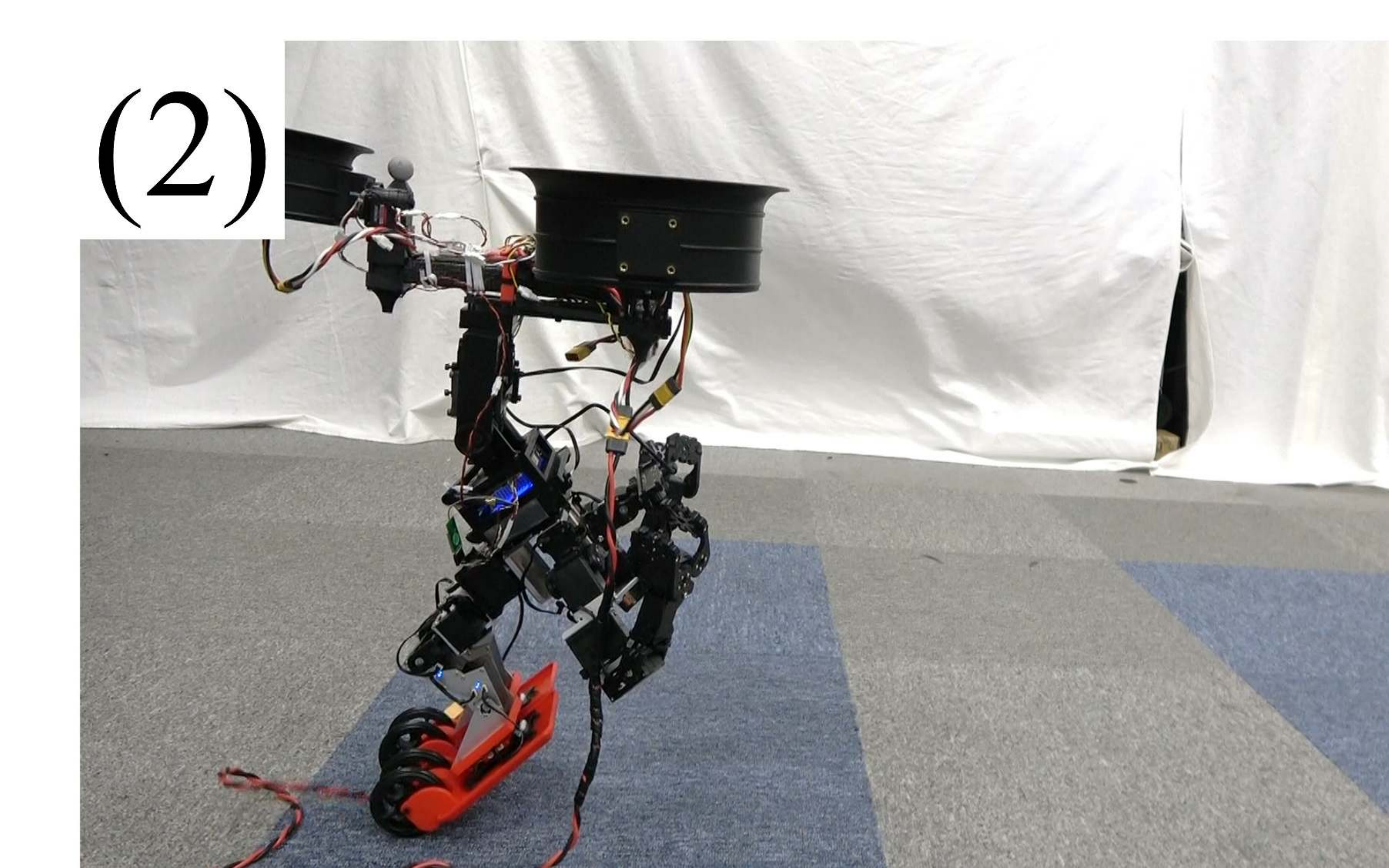}
  \includegraphics[width=0.30\columnwidth]{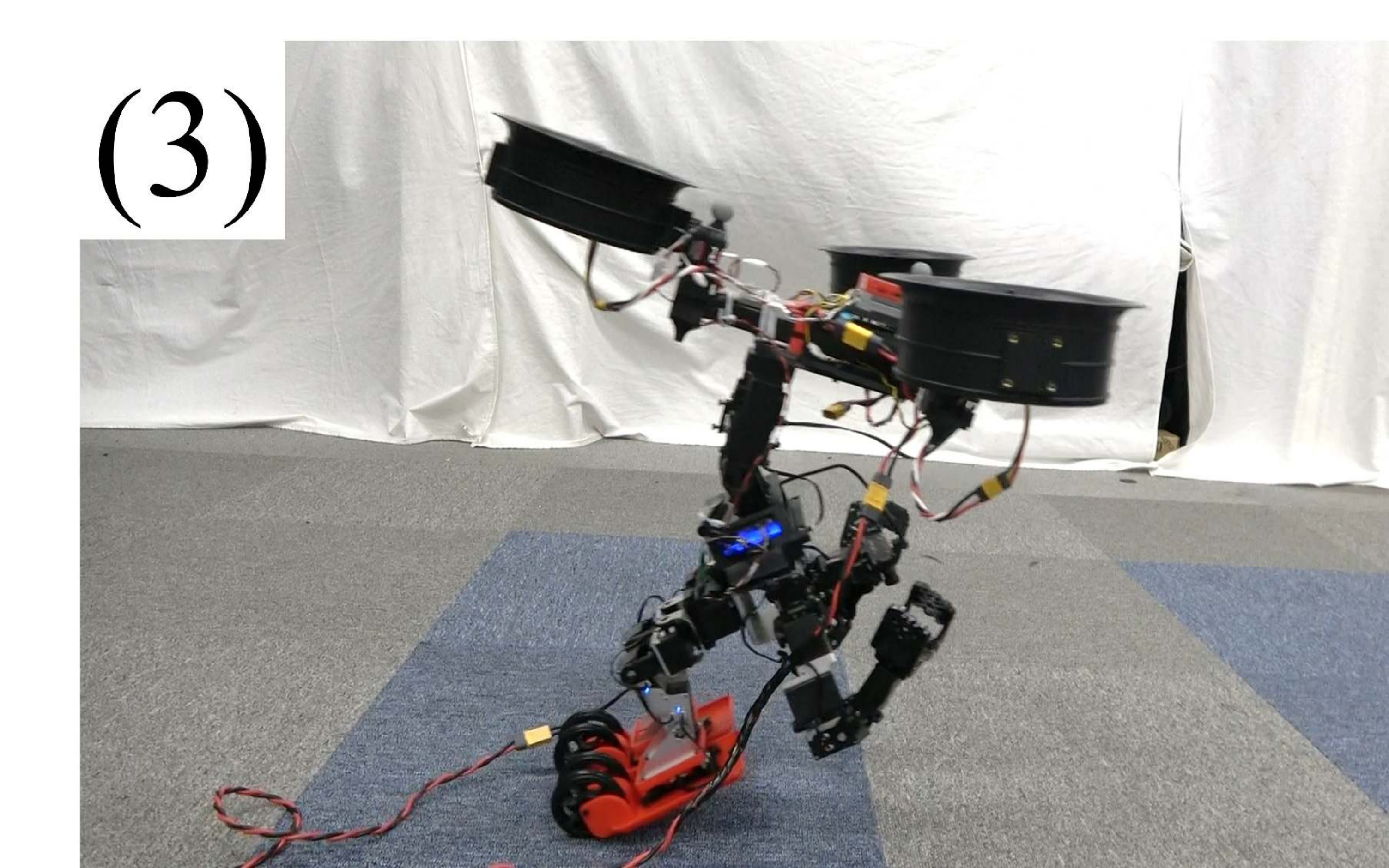}
  \vspace{-2mm}
  \caption{\textbf{Wheel locomotion experiment}. (1): Landing with foot. (2): Landing with wheel. (3): Moving forward.}
  \label{figure:wheel experiment}
\end{figure}

\begin{figure}[b]
  \centering
  \includegraphics[width=1.0\columnwidth]{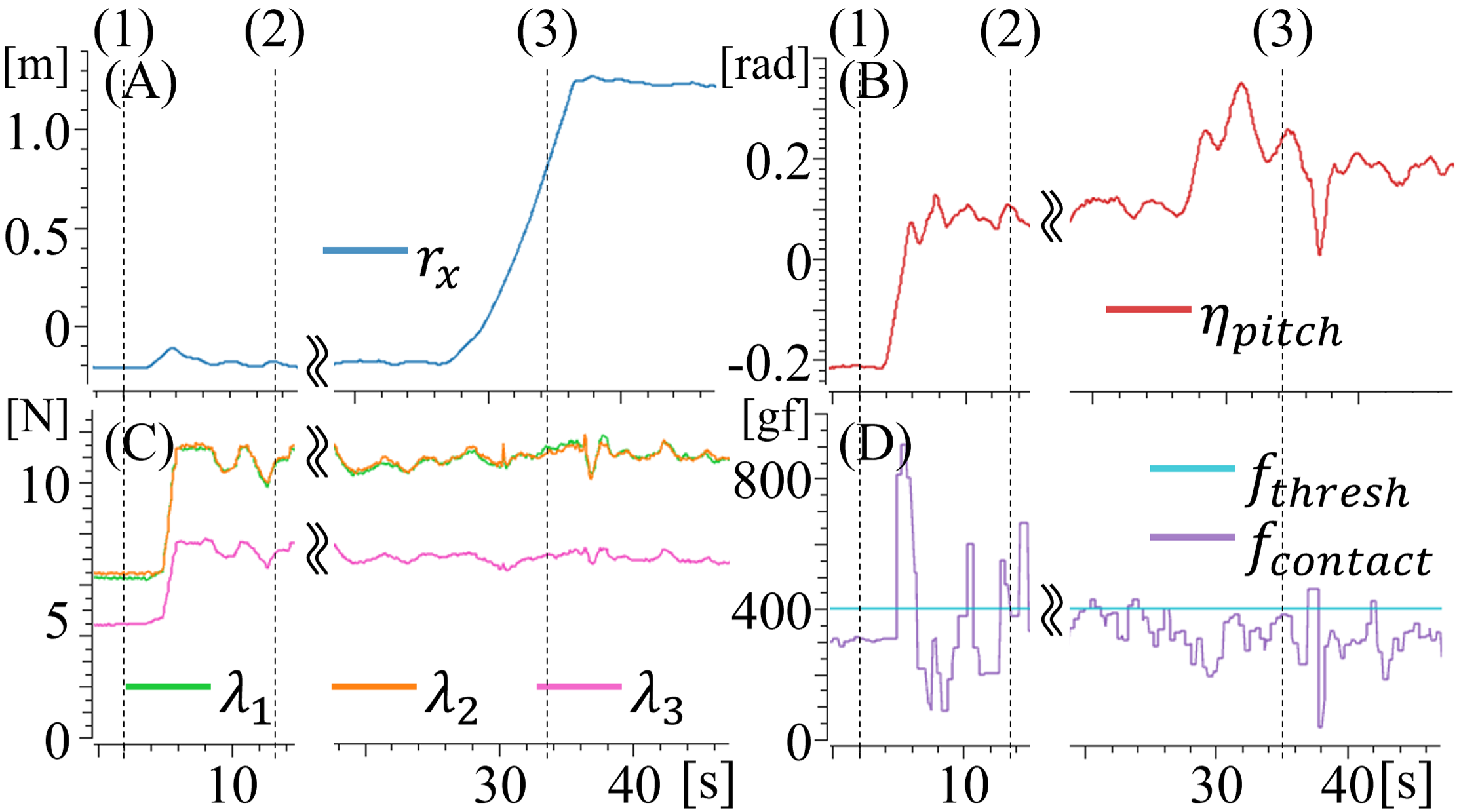}
  \vspace{-8mm}
  \caption{\textbf{Plots related to \figref{wheel experiment}}. (A): Position of flight unit. (B): Pitch angle of flight unit. (C): Thrusts. (D): Contact force and target conact forces.}
  \label{figure:running plot}
\end{figure}

\begin{figure*}[t]
  \centering
  \includegraphics[width=2.0\columnwidth]{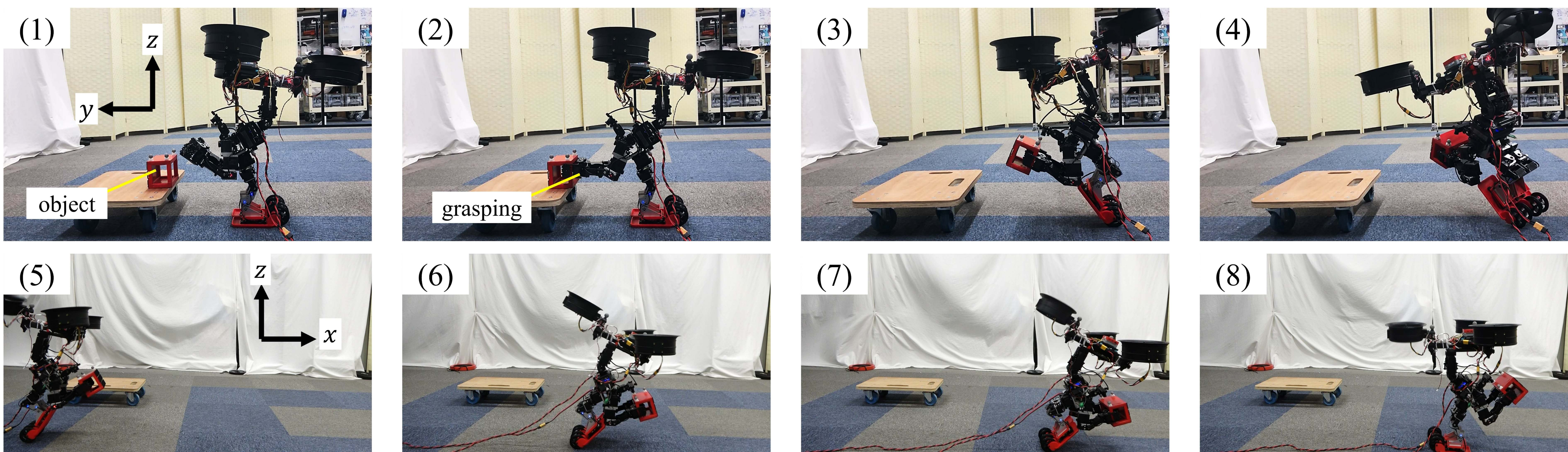}
  \vspace{-3mm}
  \caption{\textbf{Object transportation}. (1)--(2): Pick up an object. (3): Transform to wheel locomotion mode. (4): Rotate around yaw axis. (5)--(7): Move forward. (8): Transform to stand on its feet.}
  \label{figure:transportation experiment}
  \vspace{-5mm}
\end{figure*}

\begin{figure}
  \centering
  \includegraphics[width=1.0\columnwidth]{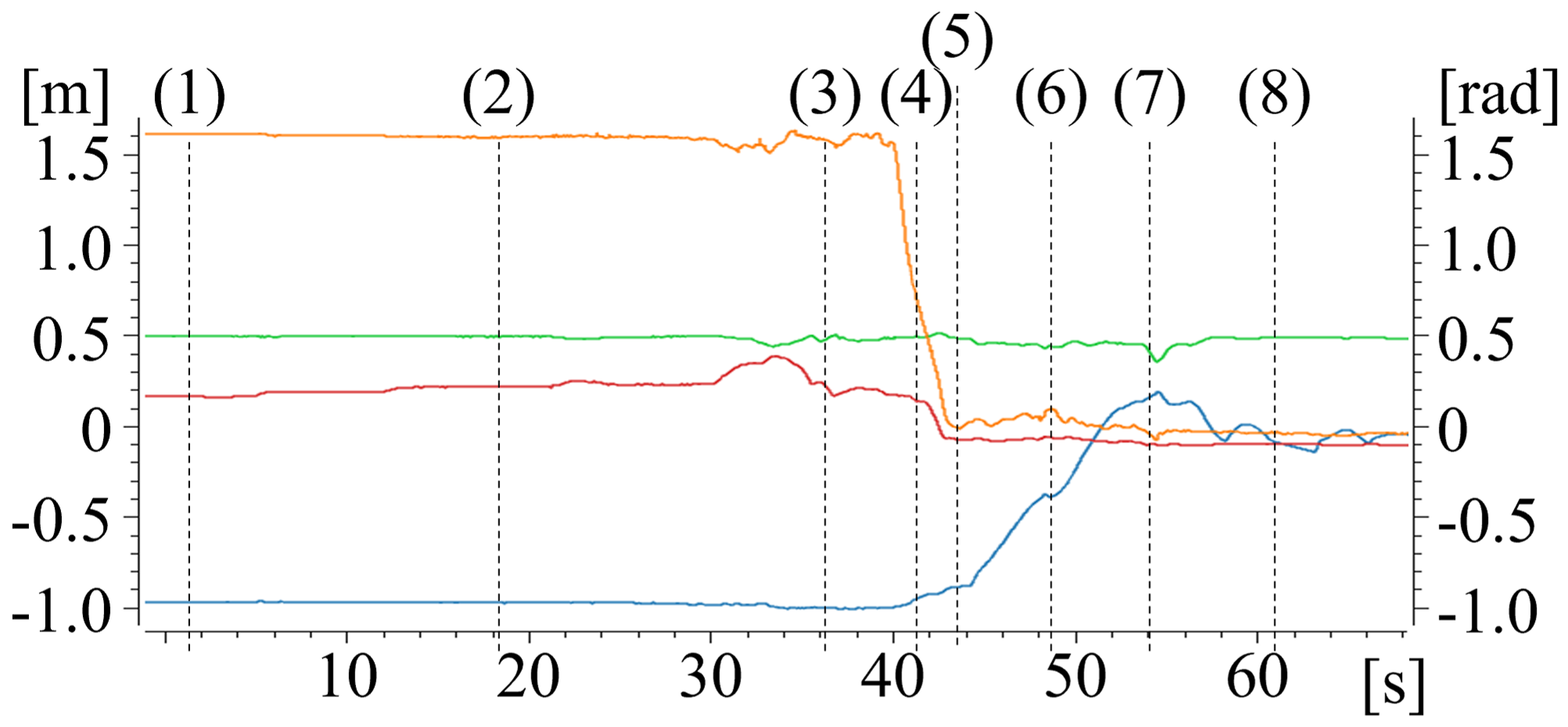}
  \vspace{-7mm}
  \caption{\textbf{Plot related to \figref{transportation experiment}.} Position and yaw angle of the flight unit in the world frame.}
  \label{figure:transportation plot}
  \vspace{-3mm}
\end{figure}



\subsubsection{Leg locomotion using thrust}
\label{section:experiment leg}
We conducted walking experiments, using the thrust control described in \secref{leg locomotion} and also without thrust.
\(\theta_{clutch}\) was the same as in aerial locomotion experiment and \( \alpha_{stable}\) in \eqref{leg clamp} was set to \(0.4 g\) \si{m/s^2} for this prototype robot. 
Without thrust control, the walking motion was unstable and the robot fell down.
As shown in \figref{walking experiment}, walking motion was achieved without falling by using thrust to stabilize motion.
For a body weight of 3.5 \si{kg}, the sum of thrust exerted by rotors was approximately 15 \si{N}.
As described in \secref{leg locomotion}, the target position was updated  with the current position.
Generally, if the target position was updated in such a way, target linear acceleration becomes constant and translational motion is uniformly accelerated.
However, in this walking experiment, the foot can receive friction from the ground, and walking motion is achieved.
\figref{walking plot} is a plot of position and attitude.
The target roll and yaw angles are updated based on a robot model and and is followed by an actual attitude.
Walking velocity was 0.003 \si{m/s}, which is relatively slow.
The main reason for this relativly slow speed is that the flight unit for the prototype in this work utilized a ducted fan module, which is safe but heavy, and the links and servo motors in legs are weak; hence, the load of the leg joints are large and it was difficult for the robot to raise its legs.




\subsubsection{Wheel locomotion}
To evaluate the feasibility of multimodal locomotion on the ground, an experiment including transformation and wheel locomotion was conducted.
Initially, the feet were completely contacted with the ground.
Then start thrust control and drive the leg joint to stand by only wheels.
At the beginning of the transformation, the force feedback control described in \secref{wheel locomotion} commenced.
In this experiment, \(\theta_{clutch}\) was 35\degree{}. The \(\beta_{stable}\) in \eqref{wheel clamp} was \(0.75 g\) \si{m/s^2}, which is larger than that of legged locomotion for more stable contact and falling avoidance.

\figref{running plot} presents the position, pitch angle of the flight unit, and values on force sensors.
In this work, we employed a fully-actuated trirotor that can control position and attitude independently; however, during wheel locomotion, wheels receive reaction force from the ground, the flight unit moves forward, and the pitch angle of the body increased.
The maximum pitch angle during moving forward was approximately 0.3 \si{rad}.
The RMS of error between the target and actual contact forces was 81.9 \si{gf}. This RMS was calculated for the data while it was stable after the transition.
Foot sensors for feedback control were deployed to the heel. So during the transition (6--8 \si{s} in \figref{running plot} (D)), the contact force increased.
The use of low rigidity link and weak servo motors in this prototype robot caused deflection and this is the reason of the error of contact force and target contact force.
The maximum velocity of wheel locomotion was 0.4 \si{m/s} and the sum of thrust exerted by rotors was approximately 28 \si{N}.

Wheeled locomotion was faster than legged locomotion; however, the amount of thrust exerted by rotors was larger.



\subsection{Object Transportation and Manipulation Experiments}
We conducted object transportation and manipulation experiments to demonstrate versatility of the proposed robot platform.
This experiment consists of following two parts.
\subsubsection{Object transportation}
Object transportation is shown in \figref{transportation experiment}.
Dual arms and grippers were used to pick up a rectangular object.
The trajectory of arm joints were generated by inverse kinematics and given object's position.
After picking up an object, the robot transformed to wheeled locomotion mode and steered around the yaw axis.
Then, the robot moved forward by wheels and transformed to stand on its feet.
In this experiment, the transitions of the state were operated manually according to the position of the robot.

\figref{transportation plot} shows the position and yaw angle of the flight unit in the world frame.
In this experiment, we validated the manipulation ability of our proposed robot and combined it with locomotion ability.

\subsubsection{Aerial manipulation}
\begin{figure}
    \centering
    \includegraphics[width=1.0\columnwidth]{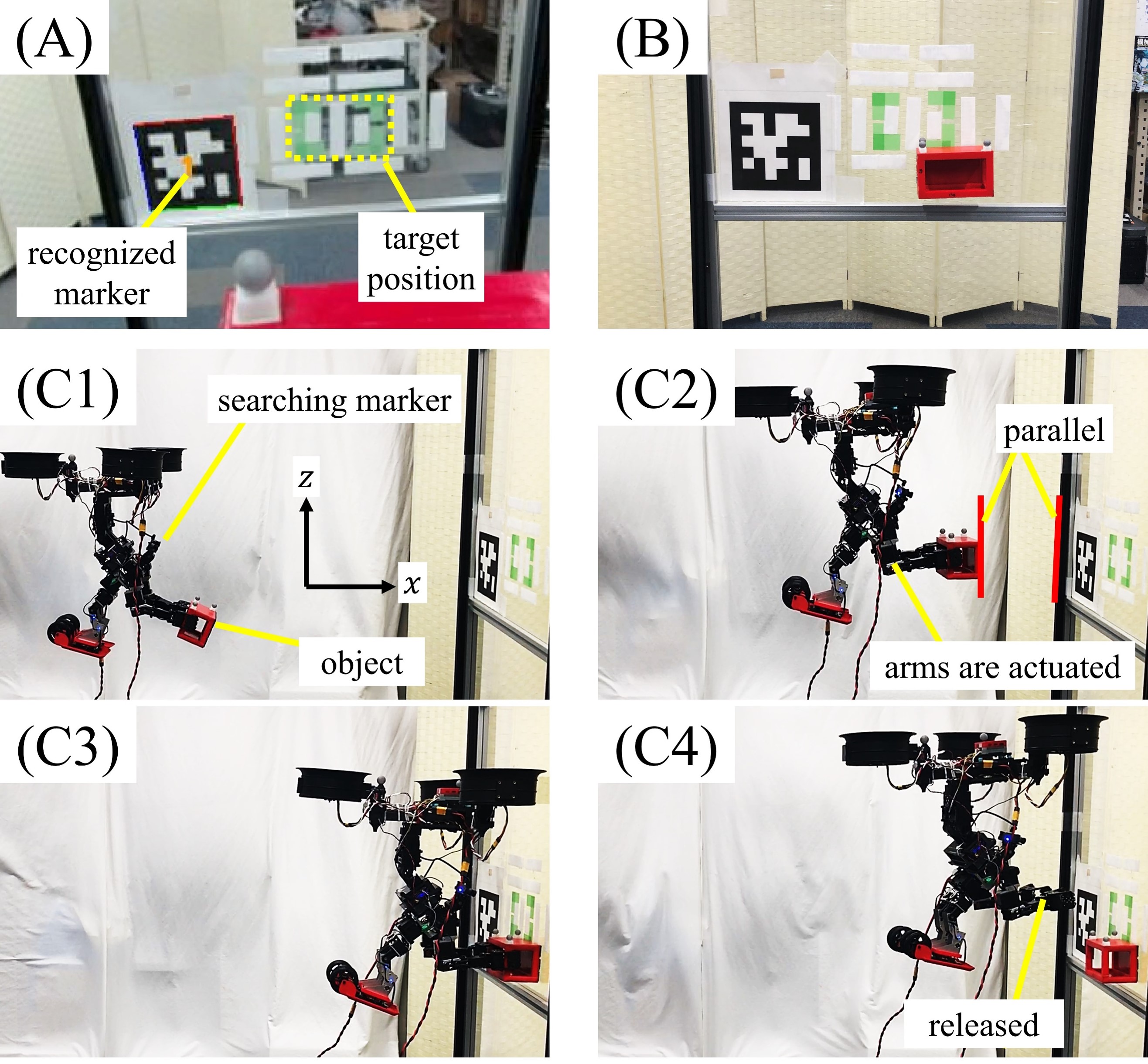}
    \vspace{-7mm}
    \caption{\textbf{Aerial manipulation}. (A): Experimental environment from vision view and recognition result. (B): Experimental result. (C1): Start searching a marker. (C2): Approaching target and  adjusting grasped object's pitch angle. (C3): Detecting contact to the wall. (C4): Release the object and leave from target.}
    \label{figure:manipulation experiment}
  \end{figure}

The aerial manipulation is shown in \figref{manipulation experiment} (C1)--(C4).
A stereo camera module mounted on the head of KHR-W was employed to capture the marker and set target position.
A rectangular object grasped by arms was attached to a target position on a wall surface by pushing it.
While approaching the target, the robot moved its arm joints so that the attitude of the grasped object, obtained by motion capture, was horizontal to the ground.
Contact to the wall was detected when the time variation of the position was below a threshold value.
After contacting for 3 \si{s}, the robot released the object and moved away from the wall.

\begin{figure}
      \centering
      \includegraphics[width=1.0\columnwidth]{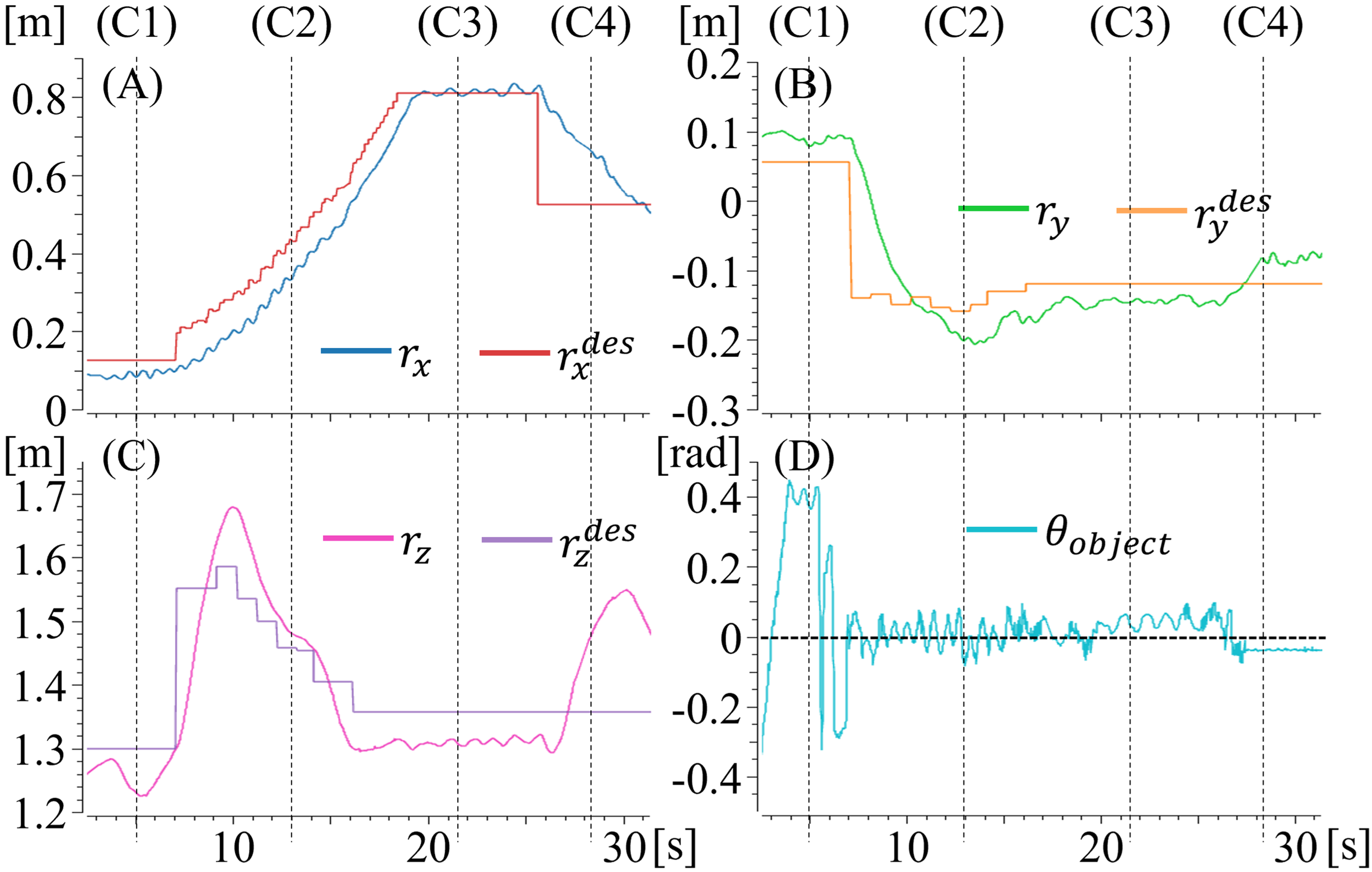}
      \vspace{-8mm}
      \caption{\textbf{Plots related to \figref{manipulation experiment}}. (A): Position of x. (B): Position of y. (C): Position of z. (D): Pitch angle of a grasped object.}
      \label{figure:manipulation plot}
      \vspace{-3mm}
    \end{figure}

\figref{manipulation experiment} (A) and (B) show the experimental environment and recognition result of the marker, a result of the experiment, respectively.
\figref{manipulation plot} shows the position data and pitch angle of the grasped object in this experiment. 
The errors of the attached and target position are smaller than 0.15 \si{m}. 
The position error remained because it contacted the wall before the y and z position control converged.
By the feedback control of dual arms using the pitch angle of the grasped object, the adhesive surface and wall became parallel and this worked effectively for stable contact motion.
As these result show, the image recognition and manipulation abilities of the humanoid could work in the air.

\section{CONCLUSIONS AND FUTURE WORK}
\label{section:conclusion}
In this work, we proposed a humanoid robot equipped with wheels and a flight unit to achieve rapid terrestrial and aerial locomotion abilities with a humanoid.
We developed a clutch module used to establish a connection between the flight unit and humanoid and enable optimal configuration for multiple modes of locomotion.
An integrated control framework for aerial, legged, and wheeled locomotions was proposed.
Then, we developed a flying humanoid robot and conducted experiments to evaluate our proposal.
In locomotion experiments, aerial, legged, and wheeled locomotions were achieved.
Through an object transportation and manipulation experiments, we demonstrated the manipulation abilities of the proposed humanoid robot worked in multiple domains.

In the future, it is necessary for the robot to determine a suitable locomotion mode by itself.
By recognizing obstacles or walls using the field of image, it will be possible to determine the appropriate locomotion mode.
The optimized full-body design of the humanoid with flight ability will be required to realize smooth locomotion and manipulation in multiple domains.
Regarding flight control, some assumptions were applied in this work to control the robot as a single rigid body.
Considering the joint motions is necessary to control as multilink robot for dynamic motion.
Besides, the manipulation experiment in this work was a simple task.
A complex task planner and the use of legs as end-effectors will be required to perform more precise and forcefull aerial manipulation tasks.



\addtolength{\textheight}{-12cm}   


\bibliographystyle{junsrt}
\bibliography{main}

\begin{thebibliography}{10}

\bibitem{Kaneko2004hrp2}
Kenji Kaneko, et~al.
\newblock Humanoid robot hrp-2.
\newblock {\em Proceedings - IEEE International Conference on Robotics and
  Automation}, Vol. 2004, pp. 1083--1090, 2004.

\bibitem{JAXON:Kojima:Humanoids2015}
Kunio Kojima, et~al.
\newblock Development of life-sized high-power humanoid robot {JAXON} for
  real-world use.
\newblock In {\em Proceedings of the 2015 IEEE-RAS International Conference on
  Humanoid Robots (Humanoids 2015)}, pp. 838--843, November 2015.

\bibitem{UnifiedBalance:Kojio:IROS2019}
Yuta Kojio, et~al.
\newblock Unified balance control for biped robots including modification of
  footsteps with angular momentum and falling detection based on capturability.
\newblock In {\em Proceedings of The 2019 IEEE/RSJ International Conference on
  Intelligent Robots and Systems}, pp. 497--504, nov 2019.

\bibitem{Kajita2003preview}
Shuuji Kajita, et~al.
\newblock Biped walking pattern generation by using preview control of
  zero-moment point.
\newblock {\em Proceedings - IEEE International Conference on Robotics and
  Automation}, Vol.~2, pp. 1620--1626, 2003.

\bibitem{Kajita2010invpend}
Shuuji Kajita, et~al.
\newblock Biped walking stabilization based on linear inverted pendulum
  tracking.
\newblock {\em IEEE/RSJ 2010 International Conference on Intelligent Robots and
  Systems, IROS 2010 - Conference Proceedings}, pp. 4489--4496, 2010.

\bibitem{Jung2018hubo}
Taejin Jung, et~al.
\newblock Development of the humanoid disaster response platform drc-hubo+.
\newblock {\em IEEE Transactions on Robotics}, Vol.~34, pp. 1--17, 2 2018.

\bibitem{Haynes2017chimp}
G.~Clark Haynes, et~al.
\newblock Developing a robust disaster response robot.
\newblock {\em Journal of Field Robotics}, Vol.~34, pp. 281--304, 3 2017.

\bibitem{Page2014uavugv}
Jared~R. Page, et~al.
\newblock The quadroller: Modeling of a uav/ugv hybrid quadrotor.
\newblock pp. 4834--4841. Institute of Electrical and Electronics Engineers
  Inc., 10 2014.

\bibitem{huang2017jet}
Zhifeng Huang, et~al.
\newblock Jet-hr1: Two-dimensional bipedal robot step over large obstacle based
  on a ducted-fan propulsion system.
\newblock In {\em 2017 IEEE-RAS 17th International Conference on Humanoid
  Robotics (Humanoids)}, pp. 406--411. IEEE, 2017.

\bibitem{kim2021bipedal}
Kyunam Kim, et~al.
\newblock A bipedal walking robot that can fly, slackline, and skateboard.
\newblock {\em Science Robotics}, Vol.~6, No.~59, p. eabf8136, 2021.

\bibitem{anzai2021design}
Tomoki Anzai, et~al.
\newblock Design and development of a flying humanoid robot platform with
  bi-copter flight unit.
\newblock In {\em 2020 IEEE-RAS 20th International Conference on Humanoid
  Robots (Humanoids)}, pp. 69--75. IEEE, 2021.

\bibitem{DRAGON:Chou:ICRA2018_RAL}
M.~Zhao, et~al.
\newblock Design, modeling, and control of an aerial robot dragon: A
  dual-rotor-embedded multilink robot with the ability of
  multi-degree-of-freedom aerial transformation.
\newblock {\em IEEE Robotics and Automation Letters}, Vol.~3, No.~2, pp.
  1176--1183, April 2018.

\bibitem{DroneDoorOpening:Sugito:ICRA2022}
Nobuki Sugito, et~al.
\newblock Aerial manipulation using contact with the environment by thrust
  vectorable multilinked aerial robot.
\newblock In {\em Proceedings of The 2022 IEEE International Conference on
  Robotics and Automation}, pp. 54--60, May 2022.

\bibitem{Qin2020bicopter}
Youming Qin, et~al.
\newblock Gemini: A compact yet efficient bi-copter uav for indoor
  applications.
\newblock {\em IEEE Robotics and Automation Letters}, Vol.~5, pp. 3213--3220, 4
  2020.

\bibitem{Hu2018trirotor}
Junyan Hu, et~al.
\newblock An innovative tri-rotor drone and associated distributed aerial drone
  swarm control.
\newblock {\em Robotics and Autonomous Systems}, Vol. 103, pp. 162--174, 5
  2018.

\bibitem{Shi2022learning}
Fan Shi, et~al.
\newblock Learning agile hybrid whole-body motor skills for thruster-aided
  humanoid robots.
\newblock {\em IEEE International Conference on Intelligent Robots and
  Systems}, Vol. 2022-October, pp. 12986--12993, 2022.

\bibitem{zhao2023spidar}
Moju Zhao, et~al.
\newblock Design, modeling and control of a quadruped robot spidar: Spherically
  vectorable and distributed rotors assisted air-ground amphibious quadruped
  robot, 2023.

\bibitem{Park2018odar}
Sangyul Park, et~al.
\newblock Odar: Aerial manipulation platform enabling omnidirectional wrench
  generation.
\newblock {\em IEEE/ASME Transactions on Mechatronics}, Vol.~23, pp.
  1907--1918, 8 2018.

\bibitem{TransformHumanoid:Makabe:ICRA2022}
Tasuku Makabe, et~al.
\newblock Design and development for humanoid-vehicle transformer platform with
  plastic resin structure and distributed redundant sensors.
\newblock In {\em Proceedings of The 2022 IEEE International Conference on
  Robotics and Automation}, pp. 8526--8532, May 2022.

\end{thebibliography}

\end{document}